\documentclass[11pt]{article}

\PassOptionsToPackage{numbers,compress}{natbib}
\usepackage{midea_arxiv}

\usepackage[utf8]{inputenc}
\usepackage{hyperref}
\hypersetup{colorlinks=true,linkcolor=black,citecolor=black,urlcolor=mideablue}
\usepackage{booktabs}
\usepackage{amsfonts}
\usepackage{nicefrac}
\usepackage{microtype}
\usepackage{amsmath,amssymb,amsfonts}
\usepackage{algorithm}
\usepackage{algpseudocode}
\usepackage{multirow}
\usepackage{graphicx}
\usepackage{siunitx}
\usepackage{subcaption}
\usepackage{threeparttable}
\usepackage{makecell}
\usepackage{wrapfig}

\newcommand{\best}[1]{{\bfseries #1}}
\newcommand{\second}[1]{\underline{#1}}
\newcommand{\grouprule}{\specialrule{0.10em}{0.12em}{0.30em}}

\midearightlogo{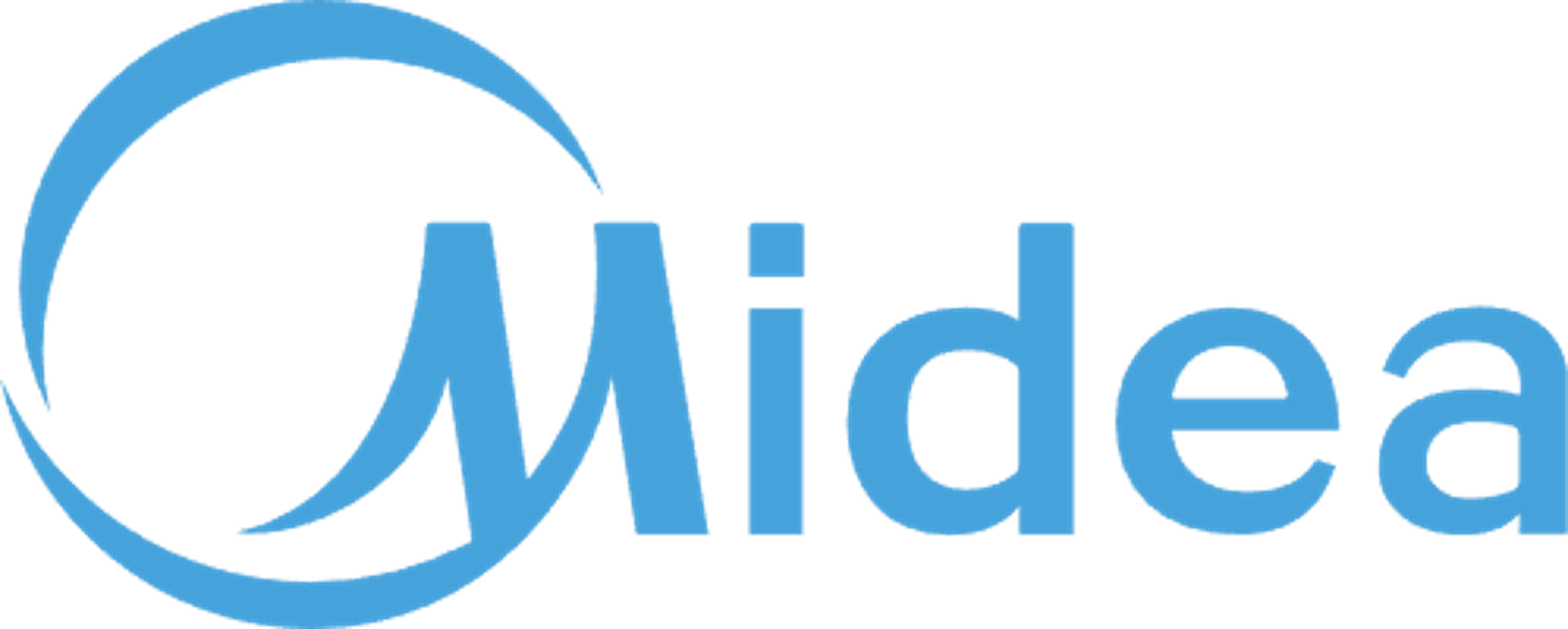}

\title{Teach-to-Reason: Competition-Guided Reasoning with a Self-Improving Teacher}

\author{%
  {\bfseries Xiao Han$^{1}$, Hao Liu$^{1,\dagger,\ddagger}$, Zhimin Bao$^{1}$, Jile Jiao$^{1}$,
  Yue Wang$^{1}$, Hui Guo$^{1}$, Xiaofeng Mou$^{1}$, Yi Xu$^{1,\dagger}$}\\[0.35em]
  {\small\ttfamily $^{1}$AIRC, Midea Group}\\[0.35em]
  {\small\ttfamily \{hanxiao36, liuhao249, baozm8, jiaojl5, wangyue184, guohui55, mouxf, xuyi42\}@midea.com}
  {\small\rmfamily $^\dagger$Corresponding Author, \quad $^\ddagger$Project Leader}
}

\reportdate{\today}
\correspondence{Hao Liu, Yi Xu}
\codeurl{https://github.com/midea-ai/Teach2Reason/}

\begin{document}

\maketitle

% ...
\begin{abstract}
Chest X-ray visual question answering (CXR VQA) requires models not only to predict correct answers, but also to produce reliable medical reasoning. However, existing reinforcement-learning-based training typically relies on answer-level rewards, which are often too coarse to improve chain-of-thought (CoT) quality and can become ineffective when group-level advantages collapse to zero. We propose \textbf{Teach-to-Reason (T2R)}, a framework that introduces comparison-based supervision into CoT optimization through a self-improving \emph{Teacher} and a competition-guided \emph{Reasoner}. As the Teacher is iteratively strengthened via self-competition, the Reasoner is optimized against progressively stronger Teacher-generated references. We further introduce a case-wise reward design that preserves the original reward-induced positive/negative partition when it is informative, and restores supervision from competition scores when the original reward signal degenerates. Experiments on multiple CXR open-ended VQA benchmarks show that T2R consistently outperforms strong baselines, indicating that comparison-based supervision, when integrated in a controlled and principled manner, provides a more effective training signal for reasoning optimization.
\end{abstract}

\section{Introduction}
Chest X-ray visual question answering (CXR VQA) aims to answer clinically relevant questions grounded in chest radiographs, and has become an important testbed for medical multimodal understanding and reasoning \cite{lau2018vqarad,liu2021slake,bae2024mimicextcxrvqa,pal2025rexvqa}. Unlike general-domain VQA, CXR VQA requires not only correct answers, but also \emph{reliable medical rationales} that justify those answers. In clinical settings, an apparently correct prediction may still rely on fragile, accidental, or clinically implausible reasoning. As a result, improving intermediate reasoning quality---rather than optimizing answer accuracy alone---is a central challenge in CXR VQA \cite{gai2025medthink,wei2022cot}.

Recent reinforcement-learning-based approaches have shown promise in improving reasoning ability in LLMs, but existing RL/RLVR-style training typically relies on \emph{answer-level} rewards, such as whether the final answer is correct or satisfies a verifiable constraint \cite{ouyang2022instructgpt,rafailov2023dpo,shao2024deepseekmath,deepseekr1,openai2024openaio1card}. Such signals are effective for optimizing final outcomes, yet they provide only weak supervision on the quality of the chain of thought (CoT) \cite{wei2022chain} itself. In particular, answer-level rewards are often too coarse to continuously shape reasoning quality, and when task rewards become identical within a group, the corresponding group-level advantage can degenerate to zero, making the training signal sparse or ineffective. A natural alternative is to directly reward CoTs using fixed rules \cite{gunjal2025rubrics} or an LLM-as-a-Judge \cite{gu2024survey}, but these approaches depend on predefined criteria and can quickly become sparse once the model learns to satisfy them \cite{zheng2023llmasjudge,yuan2024selfrewarding}.

Motivated by these limitations, we propose \textbf{Teach-to-Reason (T2R)}, a framework that introduces comparison-based supervision into CoT optimization for CXR VQA through an iteratively improved \emph{Teacher} and a competition-guided \emph{Reasoner}. In T2R, the Teacher generates reference CoTs, while the Reasoner is the model used for final VQA inference. The Teacher is improved stage by stage through \emph{self-competition}, and the Reasoner is optimized against progressively stronger Teacher-generated references. Compared with rule-based judge rewards, this design continually refreshes the supervision target as the Teacher evolves, making comparison-based supervision more adaptive and less susceptible to early saturation \cite{zelikman2022star,madaan2023selfrefine,shinn2023reflexion}.

A second key ingredient of T2R is its reward design. We do not want additional CoT supervision to overwrite or destabilize the original positive/negative partition induced by the task reward within each GRPO group. Instead, T2R follows a case-wise principle: when the task reward already induces a meaningful partition, we preserve it and use competition scores only to refine the relative ordering within each side of the partition; when the task reward becomes completely uninformative and the group-level advantage degenerates to zero, we use competition scores to recover the partition and directly construct the final training signal. In this way, T2R preserves the original reward-induced preference whenever it is informative, while restoring supervision when the original signal collapses. Thus, T2R does not replace the task reward with competition scores; rather, it augments the original partition with finer-grained and more persistent supervision.

We evaluate T2R on Qwen3-VL-Instruct models at the 2B and 4B scales \cite{bai2025qwen3vl} across multiple CXR open-ended VQA benchmarks. The results show consistent improvements over strong baselines. In summary, our main contributions are as follows:

\begin{itemize}
    \item We propose \textbf{Teach-to-Reason (T2R)}, a training framework for CXR VQA that combines \emph{iterative Teacher self-competition} with \emph{competition-guided Reasoner optimization} to improve CoT quality.
    \item We introduce a case-wise reward design that preserves the original positive/negative partition induced by task rewards within each group, uses competition scores only for within-partition refinement, and recovers supervision when the task reward becomes degenerate.
    \item Extensive experiments on multiple CXR open-ended VQA benchmarks demonstrate that T2R outperforms strong baselines, including RLVR, Frozen Teacher, and Judge Reward, and is accompanied by denser and more persistent comparison-based supervision during training.
\end{itemize}

\begin{figure*}[t]
    \centering
    \includegraphics[width=\textwidth]{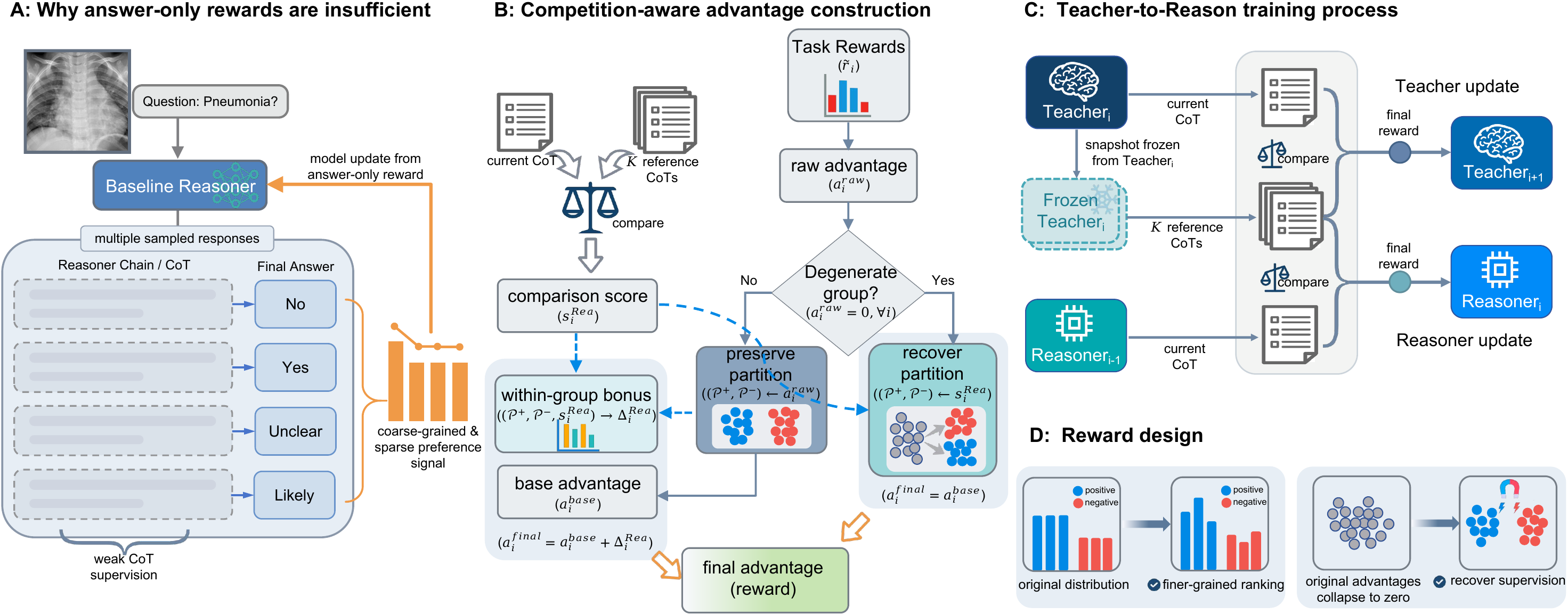}
    \caption{Overview of Teach-to-Reason (T2R). 
(A) Answer-only rewards provide only coarse and sparse supervision, making them insufficient for improving CoT quality in CXR VQA. 
(B) T2R constructs competition-aware advantages by preserving the original group partition when task rewards are informative, and recovering supervision from competition scores when the group advantage collapses to zero. 
(C) During training, the Teacher is iteratively improved via self-competition, and the Reasoner is optimized against progressively stronger Teacher-generated reference CoTs. 
(D) This yields denser and more informative training signals for CoT optimization.}
    \label{fig:overview}
\end{figure*}

\section{Related Work}

\paragraph{CXR and Medical VQA.}
Medical visual question answering has been studied through benchmarks such as VQA-RAD \cite{lau2018vqarad}, VQA-Med \cite{abacha2019vqamed}, PathVQA \cite{he2020pathvqa}, and SLAKE \cite{liu2021slake}, and more recently through larger chest-radiology resources such as MIMIC-Ext-MIMIC-CXR-VQA \cite{bae2024mimicextcxrvqa} and RexVQA \cite{pal2025rexvqa}. Prior work has also explored stronger medical VQA architectures, multimodal pretraining, and medical multimodal foundation models, including MMBERT \cite{khare2021mmbert}, M$^3$AE \cite{chen2022m3ae}, BiomedCLIP \cite{zhang2023biomedclip}, LLaVA-Med \cite{li2023llavamed}, Med-Flamingo \cite{moor2023medflamingo}, and MedGemma \cite{sellergren2025medgemma}. Together, these efforts have substantially advanced medical VQA benchmarks, model architectures, and multimodal medical modeling \cite{lin2023medvqasurvey}. However, they have mainly focused on answer prediction, dataset construction, or general medical multimodal modeling, while the explicit optimization of chain-of-thought (CoT) quality in CXR VQA remains relatively underexplored.

\paragraph{Reasoning Optimization with RL and Judge-Based Supervision.}
Recent post-training methods have improved reasoning through preference optimization and RL/RLVR-style training \cite{ouyang2022instructgpt,rafailov2023dpo,shao2024deepseekmath,deepseekr1,openai2024openaio1card}. Yet these approaches often rely on answer- or outcome-level supervision, which is typically too coarse to shape intermediate CoT quality \cite{wei2022cot,uesato2022processoutcome,lightman2023letsverify}. A more direct line of work supervises reasoning with rubrics \cite{gunjal2025rubrics,arora2025healthbench,huang2025reinforcement,shao2025dr,zhang2025chasing,zhou2025breaking}, LLM-as-a-Judge signals \cite{zheng2023llmasjudge,gu2024survey}, or self-rewarding schemes \cite{yuan2024selfrewarding}. However, such methods still assign scores to individual CoTs through predefined criteria or judge-based scoring, which may encourage criterion satisfaction but can provide limited signal for continual improvement. We instead replace direct CoT scoring with pairwise comparison, using it as a fine-grained supervision signal that can sustain informative optimization pressure throughout training.

\paragraph{Self-Evolving Supervision for Reasoning.}
Recent work has explored how language models can improve with limited external annotation by turning their own generations, feedback, or interactions into learning signals. This includes bootstrapping from self-generated rationales, critiques, and reflections \cite{zelikman2022star,madaan2023selfrefine,shinn2023reflexion}, learning from model-generated rewards or self-distilled feedback \cite{yuan2024selfrewarding,hubotter2026sdpo}, and constructing new tasks, preferences, or curricula through self-play, role-based competition, or environment-grounded interaction \cite{chen2024selfplayfinetuning,fang2025serl,zhao2025absolutezero,spice2025,searchselfplay2025,alignmentwaltz2025,he2025visplay,gao2025surveyselfevolvingagents}. These works suggest that supervision for reasoning need not remain static, but can be generated, filtered, or strengthened during learning. T2R extends the self-evolving paradigm to continual CoT optimization through an iteratively improved \emph{Teacher}. Rather than relying on direct self-refinement or self-scoring alone, T2R enables the Teacher to strengthen itself via self-competition, and uses its progressively improved CoTs to continuously guide a separate \emph{Reasoner}. In this way, T2R uses Teacher self-evolution to drive continual improvement of the \emph{Reasoner}, with progressively stronger Teacher-generated CoTs providing an evolving source of supervision for CoT optimization.

\section{Methodology}
\label{sec:method}

\subsection{Framework and Competition Scores}

\paragraph{Model Roles.}
Our framework consists of two models: a \textbf{Reasoner} and a \textbf{Teacher}. At a high level, the Teacher provides stronger reasoning references during training, while the Reasoner is the VQA model ultimately trained for inference.

The Reasoner is the \emph{target VQA model} in our framework. Given an image-question pair $(I_i,Q_i)$, it generates a chain-of-thought (CoT) together with an answer:
\begin{equation}
(\hat C_i^{\mathrm{rea}}, \hat A_i)\sim \pi_\theta^{\mathrm{rea}}(I_i,Q_i).
\end{equation}

The Teacher is used only during training. Given the image, question, reference answer, and report context, it generates higher-quality reference CoTs:
\begin{equation}
\hat C_i^{\mathrm{tea}}\sim \pi_\phi^{\mathrm{tea}}(I_i,Q_i,A_i^*,D_i),
\end{equation}
where $A_i^*$ is the reference answer and $D_i$ denotes the radiology report or other contextual document. Rather than serving as the final VQA predictor, the Teacher provides reasoning references for the Reasoner and improves its own CoT quality through self-competition.

\paragraph{Comparison-Based Scoring.}
We use an LLM-as-a-Judge to compare the quality of two candidate reasoning chains. Given two CoTs $C_1$ and $C_2$, we define
\begin{equation}
f_{\mathrm{comp}}(C_1,C_2)=\mathbf{1}[C_1\succ C_2],
\end{equation}
where $C_1\succ C_2$ means that $C_1$ is judged better than $C_2$ in terms of reasoning correctness, internal consistency, answer alignment, and answer support. To mitigate position bias, we randomize the order of the two CoTs in the judge prompt. Additional implementation details of the pairwise CoT comparison are provided in Appendix~\ref{app:judge_cot_comparison}.

For the Reasoner, given $K$ Teacher-generated reference CoTs $\{\hat C_{i,k}^{\mathrm{tea}}\}_{k=1}^{K}$, we define the competition score as
\begin{equation}
s_i^{\mathrm{rea}}
=
\frac{1}{K}\sum_{k=1}^{K}
f_{\mathrm{comp}}(\hat C_i^{\mathrm{rea}},\hat C_{i,k}^{\mathrm{tea}}),
\qquad
s_i^{\mathrm{rea}}\in[0,1].
\end{equation}
This score is the average win rate of the current Reasoner CoT against Teacher-generated reference CoTs. 
Analogously, we define the Teacher self-competition score $s_i^{\mathrm{tea}}$ by comparing the current Teacher output against $K$ CoTs $\{\bar C_{i,k}^{\mathrm{tea}}\}_{k=1}^{K}$ sampled from a frozen Teacher snapshot.

\subsection{Reasoner Objective}

\paragraph{Design Principle.}
We use $a_i^{\mathrm{final}}$ as the group-wise training signal for the GRPO update. When the base reward is informative, it determines the positive and negative samples, while competition scores only refine the relative ordering within each subset; otherwise, we rely on competition scores to directly construct the final training signal.

\paragraph{Base Reward.}
For a GRPO group of size $G$, let $\mathcal I=\{1,\dots,G\}$.
We first define the base task reward as
\begin{equation}
\tilde r_i^{\mathrm{rea}}
=
m_i^{\mathrm{fmt}}
\bigl(1+\mathbf{1}[\hat A_i=A_i^*]\bigr),
\end{equation}
where $m_i^{\mathrm{fmt}}\in\{0,1\}$ indicates whether the output follows the required format. We then compute the raw advantage induced by the base reward:
\begin{equation}
a_i^{\mathrm{raw}}
=
\tilde r_i^{\mathrm{rea}}
-\frac{1}{G}\sum_{j=1}^{G}\tilde r_j^{\mathrm{rea}}.
\end{equation}
This naturally induces a pair of base-reward-guided subsets,
\begin{equation}
\mathcal P^+=\{i\in\mathcal I: a_i^{\mathrm{raw}}>0\},
\qquad
\mathcal P^-=\{i\in\mathcal I: a_i^{\mathrm{raw}}<0\}.
\end{equation}
We distinguish two cases accordingly. If $\mathcal P^+\cup\mathcal P^-\neq\varnothing$, then the base reward already induces a non-trivial group-wise preference, and we treat the group as non-degenerate. If $\mathcal P^+=\mathcal P^-=\varnothing$, then all raw advantages are zero and the base reward is degenerate, in which case we instead rely on competition scores to construct the final training signal.

\paragraph{Non-Degenerate Case.}
If the base reward is non-degenerate, we directly use the subsets $\mathcal P^+$ and $\mathcal P^-$ induced by $a_i^{\mathrm{raw}}$, and employ competition scores only to refine the relative ordering within each subset. Specifically, we define the subset-wise standardized patterns
\begin{equation}
p_i^+
=
\frac{
s_i^{\mathrm{rea}}-\mu_+^{\mathrm{rea}}
}{
\sigma_+^{\mathrm{rea}}
},
\quad i\in\mathcal P^+,
\qquad
p_i^-
=
\frac{
s_i^{\mathrm{rea}}-\mu_-^{\mathrm{rea}}
}{
\sigma_-^{\mathrm{rea}}
},
\quad i\in\mathcal P^-,
\end{equation}
where $\mu_+^{\mathrm{rea}},\sigma_+^{\mathrm{rea}}$ and $\mu_-^{\mathrm{rea}},\sigma_-^{\mathrm{rea}}$ are the mean and standard deviation of the competition scores within $\mathcal P^+$ and $\mathcal P^-$, respectively. Based on these patterns, we define the group-wise competition bonus as
\begin{equation}
\Delta_i^{\mathrm{rea}}
=
\begin{cases}
\eta_+^{\mathrm{rea}}\, p_i^+, & i\in\mathcal P^+,\\[2pt]
\eta_-^{\mathrm{rea}}\, p_i^-, & i\in\mathcal P^-,
\end{cases}
\end{equation}
where $\eta_+^{\mathrm{rea}}\ge 0$ and $\eta_-^{\mathrm{rea}}\ge 0$ control the bonus strength on the positive and negative sides. To preserve the sign induced by the base reward, we define
\begin{equation}
\eta_+^{\mathrm{rea}}
=
\alpha
\min_{i\in\mathcal P^+,\, p_i^+<0}
\frac{a_i^{\mathrm{raw}}}{-p_i^+},
\qquad
\eta_-^{\mathrm{rea}}
=
\alpha
\min_{i\in\mathcal P^-,\, p_i^->0}
\frac{-a_i^{\mathrm{raw}}}{p_i^-},
\end{equation}where $\alpha\in(0,1)$ controls the safety margin. The final advantage is then
\begin{equation}
a_i^{\mathrm{final}}=a_i^{\mathrm{raw}}+\Delta_i^{\mathrm{rea}}.
\end{equation}

\paragraph{Degenerate Case.}
If the base reward is degenerate, it induces no non-trivial preference within the group. In this case, we construct the partition directly from the competition scores $s_i^{\mathrm{rea}}$. For a candidate threshold $\tau$, we define
\begin{equation}
\mathcal P^-(\tau)=\{i\in\mathcal I:\, s_i^{\mathrm{rea}}\le \tau\},
\qquad
\mathcal P^+(\tau)=\{i\in\mathcal I:\, s_i^{\mathrm{rea}}>\tau\}.
\end{equation}
Among all candidate thresholds, we choose the largest feasible $\tau$ satisfying
\begin{equation}
\sum_{i\in\mathcal P^-(\tau)} s_i^{\mathrm{rea}}
<
\gamma
\sum_{i\in\mathcal P^+(\tau)} s_i^{\mathrm{rea}},
\end{equation}
where $\gamma>0$ is a hyperparameter. If no feasible threshold exists, we set the final advantage of the whole group to zero.

Otherwise, after obtaining the induced subsets $\mathcal P^+$ and $\mathcal P^-$, we reconstruct the final training signal from the competition scores. We first globally center the scores:
\begin{equation}
p_i
=
s_i^{\mathrm{rea}}-\mu^{\mathrm{rea}},
\qquad
\mu^{\mathrm{rea}}
=
\frac{1}{G}\sum_{j=1}^{G}s_j^{\mathrm{rea}}.
\end{equation}
To make the reconstructed signal consistent with the induced partition, we then shift only the positive subset:
\begin{equation}
\hat p_i=
\begin{cases}
p_i+\lambda, & i\in\mathcal P^+,\\[2pt]
p_i, & i\in\mathcal P^-.
\end{cases}
\end{equation}
Here $\lambda$ is chosen as the minimum feasible shift that preserves the positive/negative structure implied by the partition:
\begin{equation}
\lambda_{\min}
=
\max\!\left(
-\frac{G}{|\mathcal P^-|}\min_{i\in\mathcal P^+} p_i,\;
\frac{G}{|\mathcal P^+|}\max_{i\in\mathcal P^-} p_i
\right),
\qquad
\lambda=\max(\lambda_{\min},0),
\end{equation}
with the derivation deferred to Appendix \ref{app:reasoner_derivations}. We then define the final advantage as
\begin{equation}
a_i^{\mathrm{final}}
=
\hat p_i-\frac{1}{G}\sum_{j=1}^{G}\hat p_j.
\end{equation}

\subsection{Teacher Objective}

The Teacher is trained to generate higher-quality CoTs that serve as stronger reasoning references for Reasoner optimization. We therefore define its reward as
\begin{equation}
r_i^{\mathrm{tea}}
=
m_i^{\mathrm{fmt}}\bigl(1+s_i^{\mathrm{tea}}\bigr),
\end{equation}
where $m_i^{\mathrm{fmt}}\in\{0,1\}$ enforces output format validity and $s_i^{\mathrm{tea}}$ encourages improved reasoning quality through self-competition. Thus, the Teacher is optimized to maximize reasoning quality subject to valid formatting.

\subsection{Training Procedure}
\label{sec:method_training_procedure}

We initialize the Teacher and the Reasoner from the same pretrained weights, denoted by $\mathcal T_0$ and $\mathcal R_0$, respectively. We first obtain $\mathcal T_1$ by freezing $\mathcal T_0$ as a reference model $\tilde{\mathcal T}_0$ and updating $\mathcal T_0$ against $\tilde{\mathcal T}_0$ via self-competition for $K$ optimization steps.

Training then proceeds iteratively. At iteration $i$, we freeze the current Teacher $\mathcal T_i$ as $\tilde{\mathcal T}_i$ and use it as the shared reference model for both Teacher and Reasoner updates. Conditioned on the same $\tilde{\mathcal T}_i$, the Teacher is updated via self-competition to obtain $\mathcal T_{i+1}$, while the Reasoner is updated against Teacher-generated reference CoTs to obtain $\mathcal R_i$; both updates are performed for $K$ optimization steps.

This yields the paired updates
\[
\mathcal T_i \rightarrow \mathcal T_{i+1},
\qquad
\mathcal R_{i-1} \rightarrow \mathcal R_i.
\]
Repeating this procedure for $T$ iterations produces a sequence of Reasoner models $\mathcal R_1,\dots,\mathcal R_T$.

This training schedule is efficient because Teacher improvement is driven entirely by self-competition and does not depend on the current Reasoner. Consequently, Teacher and Reasoner updates are naturally decoupled through the same frozen Teacher reference. An overview of the training pipeline is shown in Figure~\ref{fig:overview}.

\section{Experiments}

\subsection{Experimental Setup}
\label{sec:experimental_setup}

\paragraph{Models and Training Data.}
We use Qwen3-VL-Instruct \cite{bai2025qwen3vl} as the base model and conduct experiments at the 2B and 4B scales. Unless otherwise specified, the Teacher and the Reasoner at each scale are initialized from the same pretrained checkpoint.

Our training data is derived from RexVQA \cite{pal2025rexvqa}, a single-choice chest X-ray VQA dataset. For training efficiency, we randomly sample 67K instances. To reduce the format gap between training and evaluation, we convert 13.4K samples (approximately 20\%) into open-ended form by removing the answer options, while keeping the remaining samples in their original single-choice format.

\paragraph{Training Details.}
Unless otherwise specified, we use a GRPO group size of 16 and a batch size of 128. For competition scoring, we set the number of Teacher-generated reference CoTs to $K=10$ per sample. The safety coefficient is set to $\alpha=0.9$, and the degenerate-case threshold is set to $\gamma=0.3$. During training, both pairwise CoT comparison and open-ended answer verification are performed using Qwen3-4B-Instruct-2507 \cite{yang2025qwen3}.

To instantiate the staged training framework, we partition the 67K training set into four disjoint splits: split 0 contains one-half of the data, while splits 1--3 each contain one-sixth. Let $\mathcal T_0$ and $\mathcal R_{\mathrm{init}}$ denote the Teacher and Reasoner initialized from the same pretrained model. We first train $\mathcal R_{\mathrm{init}}$ with RLVR on split 0 for one epoch to obtain $\mathcal R_0$. Starting from $\mathcal T_0$, we then train the Teacher via self-competition on splits 1--3 to obtain $\mathcal T_1,\mathcal T_2,\mathcal T_3$. Starting from $\mathcal R_0$, the Reasoner is subsequently optimized stage by stage on splits 1--3 against these progressively improved Teacher models, yielding $\mathcal R_1,\mathcal R_2,\mathcal R_3$.
Additional hyperparameter settings and prompt details are provided in Appendix \ref{app:training_details}.

\paragraph{Baselines.}
In addition to \textbf{Teach-to-Reason (T2R)}, we compare against three baselines. All Reasoner variants are initialized from the same $\mathcal R_0$ and are further trained on splits 1--3 with matched GRPO hyperparameters and training budgets, so that all methods use the same overall data budget over splits 0--3.

\textbf{RLVR Only} starts from $\mathcal R_0$ and continues RLVR on splits 1--3 without additional CoT supervision. \textbf{Frozen Teacher} starts from $\mathcal R_0$ and optimizes the Reasoner on splits 1--3 against Teacher-generated reference CoTs, while keeping the Teacher fixed as $\mathcal T_0$ throughout. \textbf{Judge Reward} replaces pairwise comparison with a binary CoT-quality signal assigned by an LLM judge according to predefined acceptability criteria. Full details of the judging criteria, prompt design, and reward formulation for \textbf{Judge Reward} are provided in Appendix~\ref{app:judge_reasoning_quality}.

\paragraph{Evaluation.}
We evaluate our method on six CXR open-ended VQA benchmarks: Medical-CXR-VQA \cite{hu2025medicalcxrvqa}, VQA-RAD \cite{lau2018vqarad}, SLAKE \cite{liu2021slake}, MIMIC-CXR-VQA \cite{bae2023ehrxqa}, Covid19\_heywhale \cite{chowdhury2020can}, and Chest\_X-Ray\_PA \cite{asraf2021covid19}. To ensure a consistent evaluation protocol, we retain only the \textbf{CXR open-ended VQA} subset from each dataset. 
For answer evaluation, we use Qwen3-235B-A22B \cite{yang2025qwen3} as the judge. Given the question, ground-truth answer, and model prediction, the judge determines whether the prediction is correct. To improve robustness, each prediction is judged five times independently, and the final correctness label is determined by majority vote. The evaluation prompt and judge hyperparameters are provided in Appendix~\ref{app:judge_openended}. All evaluated VQA models decode with greedy search. Unless otherwise specified, we report accuracy on all benchmarks.

\begin{table*}[t]

\caption{Main results on six CXR open-ended VQA benchmarks. Best results within each model scale and dataset are shown in bold. Rows corresponding to Teach-to-Reason (T2R) are shaded in light gray and report different Reasoner stages.}
\vspace{10pt}

\centering
\small
\setlength{\tabcolsep}{4.5pt}
\renewcommand{\arraystretch}{0.96}

\resizebox{\textwidth}{!}{%
\begin{tabular}{
    c
    l
    *{6}{S[table-format=2.2]}
}
\toprule
\multirow{2}{*}{\textbf{Scale}} &
\multirow{2}{*}{\textbf{Method}} &
\multicolumn{6}{c}{\textbf{Datasets}} \\
\cmidrule(lr){3-8}
&
& \multicolumn{1}{c}{\textbf{Med-CXR}}
& \multicolumn{1}{c}{\textbf{VQA-RAD}}
& \multicolumn{1}{c}{\textbf{SLAKE}}
& \multicolumn{1}{c}{\textbf{MIMIC-CXR}}
& \multicolumn{1}{c}{\textbf{Covid19}}
& \multicolumn{1}{c}{\textbf{CXR-PA}} \\
\midrule

\multirow{7}{*}{\textbf{2B}}
& Base Model & 15.94 & 31.69 & 52.87 & 12.98 & 22.64 & 31.72 \\
& RLVR Only  & 16.87 & 30.18 & \second{53.92} & 11.85 & 21.69 & 20.32 \\
& Judge Reward  & \second{28.00} & 35.84 & {\best{56.02}} & 17.53 & 15.66 & 27.04 \\
& Frozen Teacher        & 21.74 & 40.37 & 52.35 & 20.88 & 31.88 & 32.20 \\
\cmidrule(lr){2-8}
\rowcolor{gray!8}
& T2R-R1                & 22.50 & \second{41.28} & 52.35 & 20.55 & \second{39.43} & 34.64 \\
\rowcolor{gray!8}
& T2R-R2                & 27.51 & 40.90 & \second{53.92} & \second{22.17} & 30.56 & \second{36.98} \\
\rowcolor{gray!8}
& T2R-R3                & {\best{31.73}} & \best{43.18} & 50.78 & {\best{23.87}} & \best{40.37} & {\best{37.57}} \\
\grouprule

\multirow{7}{*}{\textbf{4B}}
& Base Model & 21.74 & 40.37 & 58.11 & 20.45 & 26.22 & 41.22 \\
& RLVR Only             & 21.34 & 41.13 & 56.84 & 22.67 & 28.86 & 35.81 \\
& Judge Reward          & 22.38 & 41.26 & 56.54 & 23.09 & 29.24 & 36.40 \\
& Frozen Teacher        & {\best{24.41}} & 42.01 & \best{64.92} & \second{25.85} & 27.54 & 38.59 \\
\cmidrule(lr){2-8}
\rowcolor{gray!8}
& T2R-R1                & 21.27 & 38.63 & {\second{61.25}} & 19.82 & 40.56 & 42.25 \\
\rowcolor{gray!8}
& T2R-R2                & 22.87 & {\best{46.96}} & \best{64.92} & 25.61 & \best{49.43} & {\best{45.32}} \\
\rowcolor{gray!8}
& T2R-R3                & \second{23.23} & \second{42.42} & 60.73 & {\best{28.04}} & \second{41.37} & \second{43.85} \\
\bottomrule
\end{tabular}%
}

\label{tab:main-results}
\end{table*}

\vspace{20pt}

\vspace{15pt}
\begin{figure}[t]
    \centering
    \captionsetup[subfigure]{font=small, skip=2pt}

    \begin{subfigure}[t]{0.48\textwidth}
        \centering
        \includegraphics[width=0.75\linewidth]{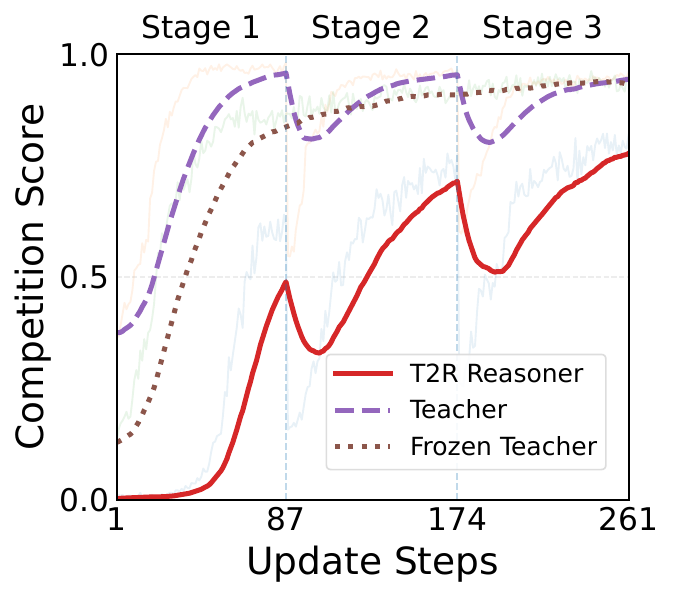}
        \caption{Competition Scores}
        \label{fig:comp-adv-a}
    \end{subfigure}
    \hfill
    \begin{subfigure}[t]{0.48\textwidth}
        \centering
        \includegraphics[width=0.75\linewidth]{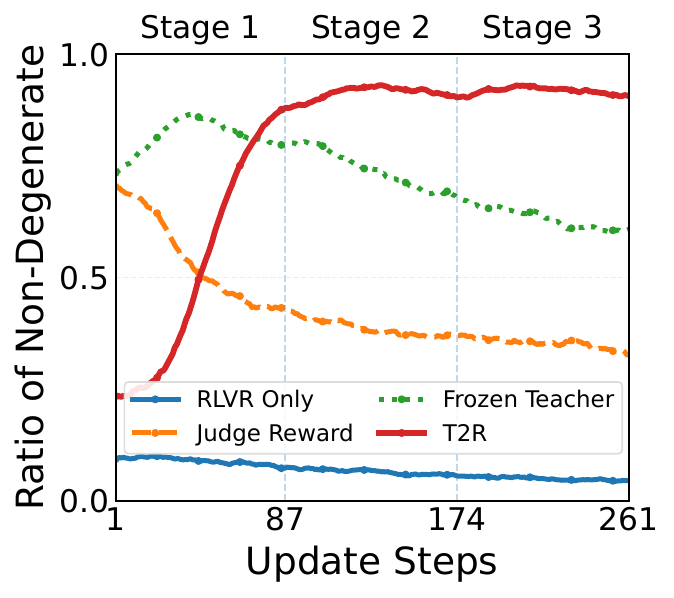}
        \caption{Non-zero Advantage}
        \label{fig:comp-adv-b}
    \end{subfigure}

    \caption{Training dynamics of (a) competition scores for the T2R Reasoner, the T2R Teacher, and Frozen Teacher, and (b) the ratio of groups with non-zero advantages for RLVR Only, Judge Reward, Frozen Teacher, and T2R.}
    \label{fig:comp-adv}
\end{figure}

\begin{figure}[t]
    \centering
    \captionsetup[subfigure]{font=small, skip=2pt}

    \begin{subfigure}[t]{0.48\textwidth}
        \centering
        \includegraphics[width=0.75\linewidth]{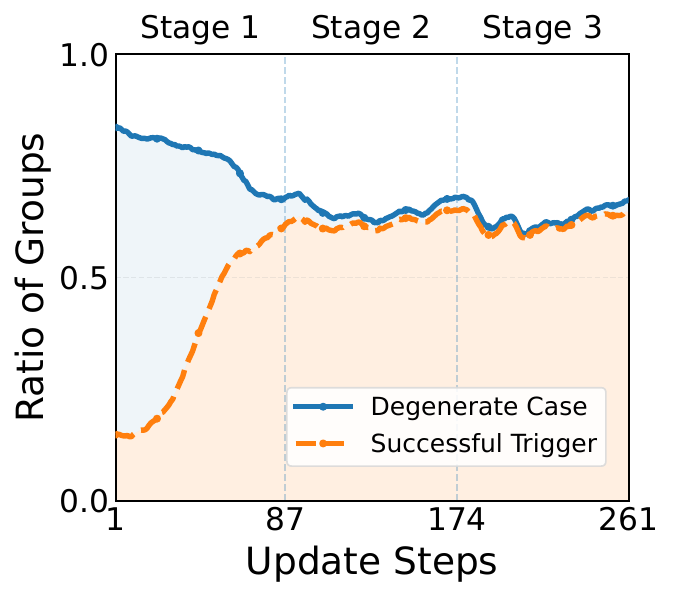}
        \caption{T2R}
        \label{fig:degenerate-a}
    \end{subfigure}
    \hfill
    \begin{subfigure}[t]{0.48\textwidth}
        \centering
        \includegraphics[width=0.75\linewidth]{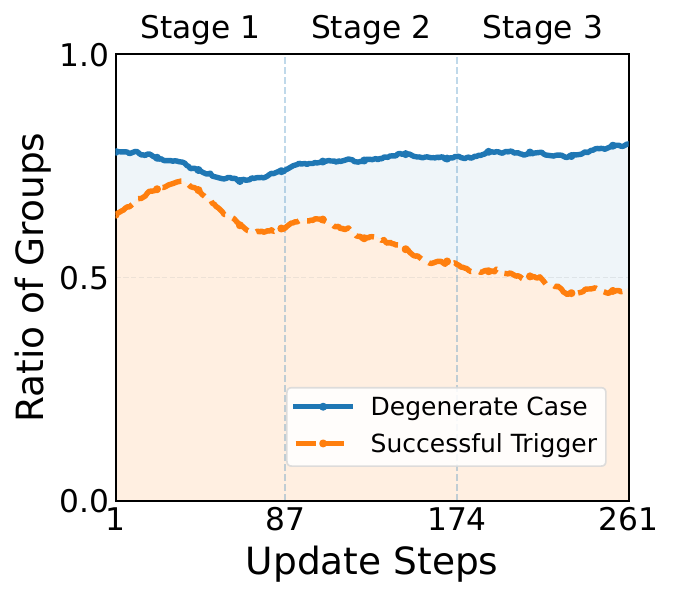}
        \caption{Frozen Teacher}
        \label{fig:degenerate-b}
    \end{subfigure}

    \caption{Degenerate-case statistics during training. (a) Ratios of degenerate groups and successfully activated degenerate-case handling groups during T2R training. (b) The same statistics for Frozen Teacher.}
    \label{fig:degenerate}
\end{figure}

\subsection{Main Results}

Table~\ref{tab:main-results} summarizes the main benchmark results at the 2B and 4B scales. For T2R, we report three Reasoner models, denoted by \textbf{T2R-R1}, \textbf{T2R-R2}, and \textbf{T2R-R3}, corresponding to $\mathcal R_1$, $\mathcal R_2$, and $\mathcal R_3$, respectively. Figure~\ref{fig:comp-adv} further shows the training dynamics of competition scores and the ratio of groups with non-zero advantages, while Figure~\ref{fig:degenerate} visualizes the activation of degenerate-case handling during training. Additional qualitative comparisons of generated CoTs are provided in Appendix~\ref{app:qualitative_cases}.

\paragraph{Overall Performance.}
T2R shows the strongest overall performance under both the 2B and 4B settings. At 2B, at least one T2R model achieves the best result on five out of six datasets, with SLAKE as the only exception, where \textbf{Judge Reward} performs best. At 4B, T2R again attains the best or tied-best result on five out of six datasets; the only exception is Medical-CXR-VQA, where \textbf{Frozen Teacher} performs best. Overall, these results indicate that comparison-guided optimization yields robust gains across diverse CXR VQA benchmarks.

\paragraph{Effect of Iterative Teacher Improvement.}
Iteratively improving the Teacher is more effective than using a frozen Teacher. As shown in Table~\ref{tab:main-results}, T2R achieves stronger best-case performance than \textbf{Frozen Teacher} under both the 2B and 4B settings. Figure~\ref{fig:comp-adv}(a) is consistent with this trend: both the T2R Teacher and the T2R Reasoner exhibit a stage-wise drop-and-recovery pattern, reflecting the introduction of progressively stronger comparison targets at each stage. In contrast, the Reasoner trained with \textbf{Frozen Teacher} gradually saturates, suggesting that a fixed Teacher provides increasingly limited comparison-based supervision over time.

\paragraph{Supervision Density During Training.}
T2R yields denser and more persistent supervision signals than \textbf{RLVR Only}, \textbf{Judge Reward}, and \textbf{Frozen Teacher}. As shown in Figure~\ref{fig:comp-adv}(b), \textbf{RLVR Only} maintains a consistently low ratio of groups with non-zero advantages, while \textbf{Judge Reward} and \textbf{Frozen Teacher} both decline over training. By contrast, T2R rises quickly and remains at the highest level in later stages, indicating that iterative Teacher improvement sustains a substantially larger fraction of trainable groups throughout optimization. This is consistent with the stronger benchmark performance of T2R in Table~\ref{tab:main-results}.

\paragraph{Impact of Degenerate-Case Handling.}
Figure~\ref{fig:degenerate} shows that degenerate-case handling is not merely activated occasionally, but remains highly engaged throughout T2R training. Under T2R, the ratio of successfully triggered groups rises quickly and stays close to the overall ratio of degenerate groups, indicating that the proposed design effectively restores supervision when the original group reward collapses. In contrast, under \textbf{Frozen Teacher}, the trigger ratio gradually decreases over time despite the continued presence of degenerate groups. This pattern highlights the importance of iterative Teacher improvement for sustaining supervision in degenerate groups.

\subsection{Ablation Study}
\begin{table}[t]
\centering
\vspace{-5pt}

\begin{minipage}[t]{0.49\linewidth}
\centering
\normalsize
\setlength{\tabcolsep}{2.8pt}
\renewcommand{\arraystretch}{1.02}

\captionof{table}{Ablation on the threshold hyperparameter $\gamma$ for degenerate-group partitioning.}
\label{tab:gamma_ablation}

\resizebox{\linewidth}{!}{%
\begin{tabular}{c*{6}{c}}
\toprule
\textbf{$\gamma$}
& \makecell{\textbf{Med-}\\\textbf{CXR}}
& \makecell{\textbf{VQA-}\\\textbf{RAD}}
& \textbf{SLAKE}
& \makecell{\textbf{MIMIC-}\\\textbf{CXR}}
& \makecell{\textbf{Covid-}\\\textbf{19}}
& \makecell{\textbf{CXR-}\\\textbf{PA}} \\
\midrule
0.1        & \best{28.11} & \second{41.00}   & \second{51.96}          & \second{22.16}   & 34.21 & \best{37.26} \\
\rowcolor{gray!8}
0.3 & \second{27.24}   & \best{41.78} & \best{52.35}   & \best{22.19} & \best{36.78}   & \second{36.39} \\
0.6        & 26.73          & 36.59          & 51.48 & 21.80          &  \second{35.33}          & 36.24 \\
0.9        & 27.01          & 38.61          & 50.25          & 21.64          & 33.86          & 35.16 \\
\bottomrule
\end{tabular}%
}
\end{minipage}
\hfill
\begin{minipage}[t]{0.49\linewidth}
\centering
\normalsize
\setlength{\tabcolsep}{2.8pt}
\renewcommand{\arraystretch}{1.02}

\captionof{table}{Effect of Teacher strength in the Frozen Teacher setting.}

\label{tab:frozen_teacher_stage}

\resizebox{\linewidth}{!}{%
\begin{tabular}{c*{6}{c}}
\toprule
\makecell{\textbf{Tea-}\\\textbf{cher}}
& \makecell{\textbf{Med-}\\\textbf{CXR}}
& \makecell{\textbf{VQA-}\\\textbf{RAD}}
& \textbf{SLAKE}
& \makecell{\textbf{MIMIC-}\\\textbf{CXR}}
& \makecell{\textbf{Covid-}\\\textbf{19}}
& \makecell{\textbf{CXR-}\\\textbf{PA}} \\
\midrule
$\mathcal{T}_0$ & \second{21.74} & \best{40.37}   & \second{52.35} & \best{20.88}   & \second{31.88}   & \second{32.20} \\
$\mathcal{T}_1$ & \best{23.78}   & \second{37.50} & 51.30          & \second{19.70} & \best{34.90} & \best{34.88} \\
$\mathcal{T}_2$ & 18.03          & 32.19          & \best{55.49}   & 13.57          & 18.30          & 25.14 \\
$\mathcal{T}_3$ & 17.52          & 33.33          & 51.30          & 13.31          & 22.64          & 24.26 \\
\bottomrule
\end{tabular}%
}
\end{minipage}

\vspace{-5pt}
\end{table}

\paragraph{Effect of $\gamma$.}

\begin{wrapfigure}{r}{0.34\linewidth}
    \vspace{-5pt}
    \centering
    \includegraphics[width=\linewidth]{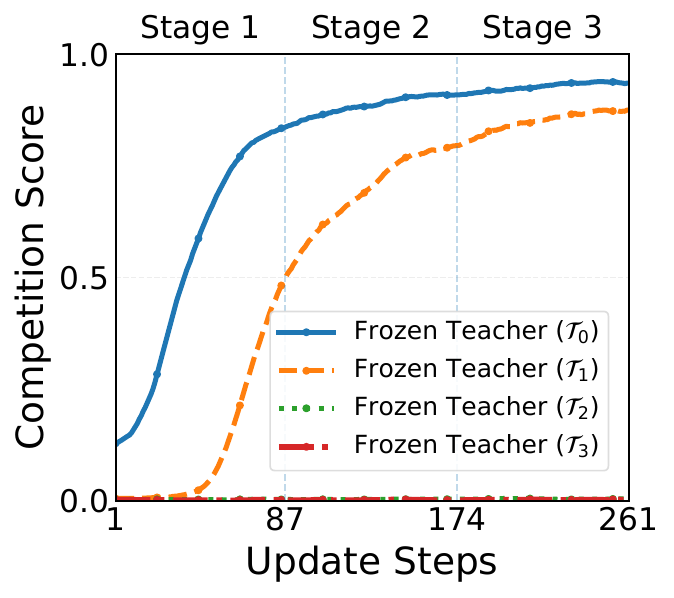}
    \vspace{-25pt}
    \caption{Competition score dynamics under different fixed Teachers in Frozen Teacher setting.}
    \label{fig:frozen-teacher-score}
\end{wrapfigure}

We study the effect of the threshold hyperparameter $\gamma$ in degenerate-case partitioning. Table~\ref{tab:gamma_ablation} reports results under the 2B setting, where each entry is the mean accuracy of the three Reasoner models $\mathcal R_1,\mathcal R_2,\mathcal R_3$ on the corresponding benchmark. Overall, $\gamma=0.3$ achieves the best trade-off, yielding the best result on four out of six datasets.

The role of $\gamma$ is to control how aggressively degenerate groups are partitioned using competition scores. Larger values make the partition more permissive, so some samples that are not clearly inferior may still be assigned to the negative subset, reducing the reliability of the reconstructed training signal. Smaller values make the partition more conservative, so some potentially useful degenerate groups may fail to trigger the reconstruction mechanism. Consistent with this intuition, overly large values ($\gamma=0.6,0.9$) perform worse overall, while an overly small value such as $\gamma=0.1$ is slightly better on a few datasets but remains less favorable. We therefore adopt $\gamma=0.3$ as the default setting.

\begin{table*}[t]
\centering
\small
\setlength{\tabcolsep}{3.8pt}
\renewcommand{\arraystretch}{0.98}

\caption{Component ablations under the 2B setting. \textit{Degenerate Handling} keeps only degenerate-case handling, \textit{Non-Degenerate Bonus} keeps only the non-degenerate bonus, and \textit{Direct Competition Reward} directly injects competition scores into the reward.}

\label{tab:ablation-components}

\resizebox{\linewidth}{!}{%
\begin{tabular}{
l
*{6}{S[table-format=2.2]}
}
\toprule
\textbf{Method}
& \multicolumn{1}{c}{\textbf{Med-CXR}}
& \multicolumn{1}{c}{\textbf{VQA-RAD}}
& \multicolumn{1}{c}{\textbf{SLAKE}}
& \multicolumn{1}{c}{\textbf{MIMIC-CXR}}
& \multicolumn{1}{c}{\textbf{Covid19}}
& \multicolumn{1}{c}{\textbf{CXR-PA}} \\
\midrule

Base Model      & 15.94 & 31.69 & \multicolumn{1}{c}{\underline{52.87}} & 12.98 & 22.64 & 31.72 \\
RLVR Only                  & 16.87 & 30.18 & \multicolumn{1}{c}{\textbf{53.92}}   & 11.85 & 21.69 & 20.32 \\
\midrule
Degenerate Handling       & \best{28.50} & \second{40.62}   & 51.30 & \second{21.20} &  \best{45.71} & \second{35.07} \\
Non-Degenerate Bonus       & 20.94 & 31.69 & \best{53.92} & 12.52 & 20.93 & 25.58 \\
Direct Competition Reward  & 26.55 & 38.74 & 51.04 & 21.14 & 33.77 & 34.63 \\
\rowcolor{gray!8}
T2R (ours)                        & \second{27.24} & \best{41.78} & 52.35 & \best{22.19} & \second{36.78} & \best{36.39} \\
\bottomrule
\end{tabular}%
}
\end{table*}

\begin{table*}[t]
\centering
\small
\setlength{\tabcolsep}{4.5pt}
\renewcommand{\arraystretch}{1.02}
\caption{CoT quality analysis via SFT distillation under the 2B setting.}
\label{tab:sft-distill}

\begin{tabular}{
l
*{6}{S[table-format=2.2]}
}
\toprule
\textbf{Method}
& \multicolumn{1}{c}{\textbf{Med-CXR}}
& \multicolumn{1}{c}{\textbf{VQA-RAD}}
& \multicolumn{1}{c}{\textbf{SLAKE}}
& \multicolumn{1}{c}{\textbf{MIMIC-CXR}}
& \multicolumn{1}{c}{\textbf{Covid19}}
& \multicolumn{1}{c}{\textbf{CXR-PA}} \\
\midrule
RLVR Only
& 13.79 & 28.03 & 51.30 & 14.28 & 23.96 & 22.07 \\
Judge Reward
& 21.40 & 33.71 & 51.83 & 16.94 & 27.73 & 26.60 \\
Frozen Teacher
& \multicolumn{1}{c}{\second{23.71}}
& \multicolumn{1}{c}{\second{36.36}}
& \multicolumn{1}{c}{\best{56.54}}
& \multicolumn{1}{c}{\second{20.99}}
& \multicolumn{1}{c}{\second{27.90}}
& \multicolumn{1}{c}{\second{32.98}} \\
\rowcolor{gray!8}
T2R-R3
& \multicolumn{1}{c}{\best{26.44}}
& \multicolumn{1}{c}{\best{39.77}}
& \multicolumn{1}{c}{\second{56.02}}
& \multicolumn{1}{c}{\best{21.28}}
& \multicolumn{1}{c}{\best{30.75}}
& \multicolumn{1}{c}{\best{33.04}} \\
\bottomrule
\end{tabular}
\end{table*}

\paragraph{Component Ablation.}
We ablate the main competition-based components of T2R in Table~\ref{tab:ablation-components}. \textbf{Degenerate Handling} retains only the competition-guided partition and supervision recovery in degenerate groups. \textbf{Non-Degenerate Bonus} retains only the sign-preserving bonus in non-degenerate groups. \textbf{Direct Competition Reward} removes the case-wise design and instead directly injects the competition score into the training reward:
\begin{equation}
\tilde r_i^{\mathrm{rea}}
=
m_i^{\mathrm{fmt}}
\bigl(1+s_i^{\mathrm{rea}}+\mathbf{1}[\hat A_i=A_i^*]\bigr).
\end{equation}
As shown in Table~\ref{tab:ablation-components}, \textbf{Degenerate Handling} yields larger gains than \textbf{Non-Degenerate Bonus} on most datasets, suggesting that a major part of T2R's improvement comes from recovering supervision in degenerate groups. By contrast, \textbf{Non-Degenerate Bonus} still improves over \textbf{RLVR Only} on most datasets, indicating that within-group refinement is helpful but not the dominant source of gain.

Meanwhile, \textbf{Direct Competition Reward}, although stronger than \textbf{Base Model} and \textbf{RLVR Only}, remains weaker than both \textbf{Degenerate Handling} and the full \textbf{T2R}. This suggests that competition scores are most effective when incorporated into the reward in a controlled and principled way, rather than being directly injected into the original reward.

\paragraph{Importance of a Comparable Teacher.}
We use the \textbf{Frozen Teacher} setting to study how Teacher strength affects Reasoner training, by fixing the Teacher as $\mathcal T_0$, $\mathcal T_1$, $\mathcal T_2$, or $\mathcal T_3$. Figure~\ref{fig:frozen-teacher-score} shows that the Reasoner's competition score increases over training when the fixed Teacher is $\mathcal T_0$ or $\mathcal T_1$, but remains near zero when it is $\mathcal T_2$ or $\mathcal T_3$. Consistently, Table~\ref{tab:frozen_teacher_stage} shows that $\mathcal T_0$ and $\mathcal T_1$ yield much better results. These observations suggest that effective comparison-based supervision requires a Teacher of comparable difficulty: if the Teacher is too strong, the comparison signal becomes overly sparse, leaving little room for the Reasoner to improve. This also motivates the stage-wise design of T2R, where the Teacher is updated after each stage so that it remains comparable to the current Reasoner.

\subsection{CoT Quality Analysis via SFT Distillation}

We further analyze the quality of the generated \textit{CoT + answer} rollouts through an SFT distillation protocol. Specifically, we randomly sample 20K inputs from the 67K training set and use \textbf{RLVR Only}, \textbf{Judge Reward}, \textbf{Frozen Teacher}, and \textbf{T2R-R3} to generate one rollout per input using greedy decoding. Each rollout set is then used to supervise the same Qwen3-VL-2B-Instruct student model for one epoch under identical initialization and identical SFT hyperparameters. The resulting students are evaluated on the same six CXR open-ended VQA benchmarks.

Although this protocol does not measure CoT quality in complete isolation, it provides a more controlled proxy than directly asking an LLM to compare CoTs, which may be affected by position bias, length bias, and judge preference. If the rollouts from one method consistently yield a stronger student under the same SFT budget, this provides evidence that its generated supervision is of higher practical quality.

Table~\ref{tab:sft-distill} shows a clear progression across the four rollout sources. \textbf{RLVR Only} yields the weakest student across the six benchmarks, suggesting that improving answers alone is insufficient for producing high-quality reasoning traces and that explicit CoT supervision is necessary. Both \textbf{Frozen Teacher} and \textbf{T2R-R3} then consistently outperform \textbf{Judge Reward}, indicating that comparison-based supervision is more effective than direct judge-based reward for improving CoT quality. Finally, among the comparison-based methods, \textbf{T2R-R3} further outperforms \textbf{Frozen Teacher} on most benchmarks, suggesting that iterative Teacher refinement further improves the quality of the generated CoTs, as reflected by the stronger student distilled from \textbf{T2R-R3} rollouts.

\section{Conclusion}

We presented \textbf{Teach-to-Reason (T2R)}, a training framework for CXR VQA in which an iteratively self-improving Teacher provides progressively stronger references for a comparison-guided Reasoner, while the original reward-induced partition is preserved whenever it is informative and supervision is restored from competition scores when the original signal degenerates. 

More broadly, T2R suggests a new training paradigm in which \emph{comparison}, rather than direct score assignment, serves as the reward signal for reinforcement learning. In this work, comparison is used only to judge which CoT is better overall, but the same idea could be extended much further by defining multiple finer-grained comparisons, each tied to a specific rule, preference, or task requirement. Moreover, this paradigm is not limited to CoT optimization: it can in principle be applied to other model outputs, such as solution strategies, plans, code, or other structured outputs, pointing toward more controllable RL-based training through customizable comparison signals.

{
    \small
    \bibliographystyle{unsrtnat}
    \bibliography{midea_arxiv}
}


\newpage

\appendix

\section{LLM-as-a-Judge Details}
\label{app:judge_details}

Unless otherwise specified, we use \texttt{Qwen3-4B-Instruct-2507} as the judge model and deploy it with \texttt{vLLM}, using \texttt{max\_model\_len=32768}, \texttt{gpu\_mem\_util=0.9}, \texttt{max\_tokens=2048}, \texttt{temperature=0.7}, and \texttt{top\_p=0.9}. Across all judging tasks, the model is required to return a valid XML response with a task-specific label in the \texttt{<result>} field. For binary yes/no judging tasks, if the XML output is malformed or the \texttt{<result>} field is invalid, we apply a unified fallback rule: if the decoded response contains the substring \texttt{>no<} in a case-insensitive manner, we assign \texttt{no}; otherwise, we assign \texttt{yes}.

\subsection{CoT Comparison for Competition Scoring}
\label{app:judge_cot_comparison}

For pairwise CoT comparison during training, the judge receives a chest X-ray report, an exam question, the official correct answer, and two candidate reasoning traces. It is asked to determine which reasoning trace is overall better, based on content quality rather than length. The comparison criteria include reasoning correctness, internal consistency, answer alignment, answer support, relevance, restraint, and clarity. The comparison prompt is shown in Figure~\ref{fig:compare_prompt}.

To mitigate position bias, the two candidate CoTs are randomly assigned to the prompt slots \texttt{\{reason\_a\}} and \texttt{\{reason\_b\}}. The judge must return exactly one winner, either \texttt{A} or \texttt{B}; tie outputs are not allowed. After decoding, the predicted label is mapped back to the original ordering of the compared CoTs before computing the comparison outcome.

For this task, malformed XML outputs or invalid \texttt{<result>} labels are discarded rather than repaired. After filtering invalid comparisons, we compute the competition score using the retained valid outputs. The Reasoner competition score is
\begin{equation}
s_i^{\mathrm{rea}}
=
\frac{1}{K}\sum_{k=1}^{K}
f_{\mathrm{comp}}(\hat C_i^{\mathrm{rea}},\hat C_{i,k}^{\mathrm{tea}}),
\qquad
s_i^{\mathrm{rea}}\in[0,1],
\end{equation}
and the Teacher self-competition score is defined analogously as
\begin{equation}
s_i^{\mathrm{tea}}
=
\frac{1}{K}\sum_{k=1}^{K}
f_{\mathrm{comp}}(\hat C_i^{\mathrm{tea}},\bar C_{i,k}^{\mathrm{tea}}),
\qquad
s_i^{\mathrm{tea}}\in[0,1].
\end{equation}


\subsection{Open-Ended Answer Judging}
\label{app:judge_openended}

For open-ended answer grading, the judge receives a radiology report, an open-ended question, a reference answer, and a predicted answer, and determines whether the prediction should be considered correct. The prompt instructs the judge to ground its decision primarily in the radiology report, while using the reference answer to identify the core medical elements required by the question. The grading prompt is shown in Figure~\ref{fig:judge_prompt}.

\paragraph{Training-time setting.}
During training, we use the default judge model and configuration described above. Each sample is judged once, and the judge is required to return a binary decision with \texttt{<result>} equal to either \texttt{yes} or \texttt{no}. When the XML output is malformed or the label is invalid, we apply the unified fallback rule so that each training sample yields a binary supervision signal.

\paragraph{Final-evaluation setting.}
At final evaluation time, we replace the default judge with a stronger model, \texttt{Qwen3-235B-A22B}, while keeping the same prompt template and per-run decoding configuration. Each sample is judged five times independently, and the final label is determined by majority vote over the five binary decisions. The same fallback rule is applied to each run before aggregation. This repeated-judging strategy stabilizes the final LLM-as-a-Judge decision and reduces evaluation variance.

\mideafullpagefigure[max width=0.96\paperwidth,max height=0.88\paperheight]{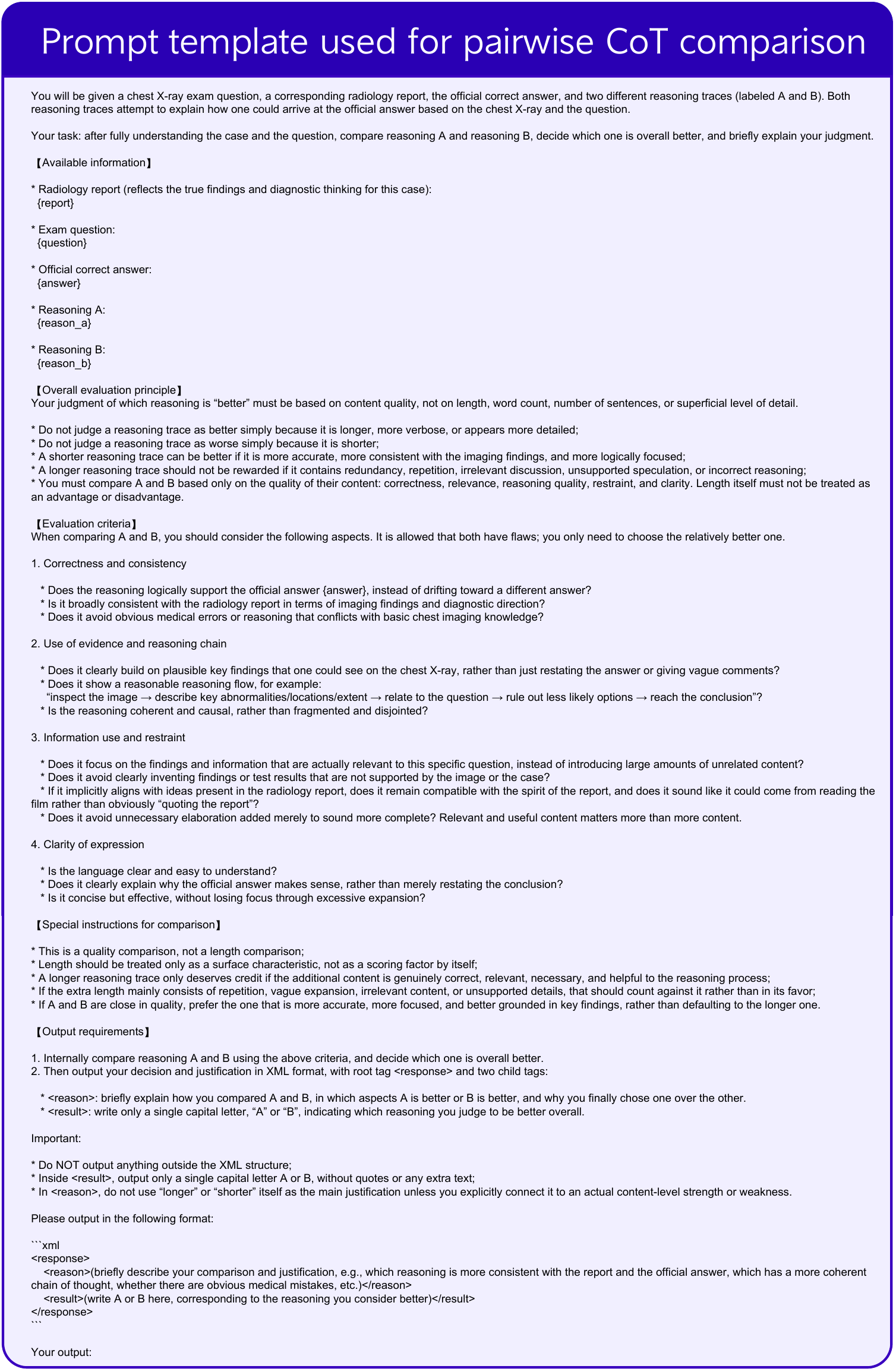}
{Prompt template used for pairwise CoT comparison in the LLM-as-a-Judge module. The two candidate reasoning traces are randomly assigned to slots A and B to reduce position bias.}
[fig:compare_prompt]

\mideafullpagefigure[max width=0.96\paperwidth,max height=0.88\paperheight]{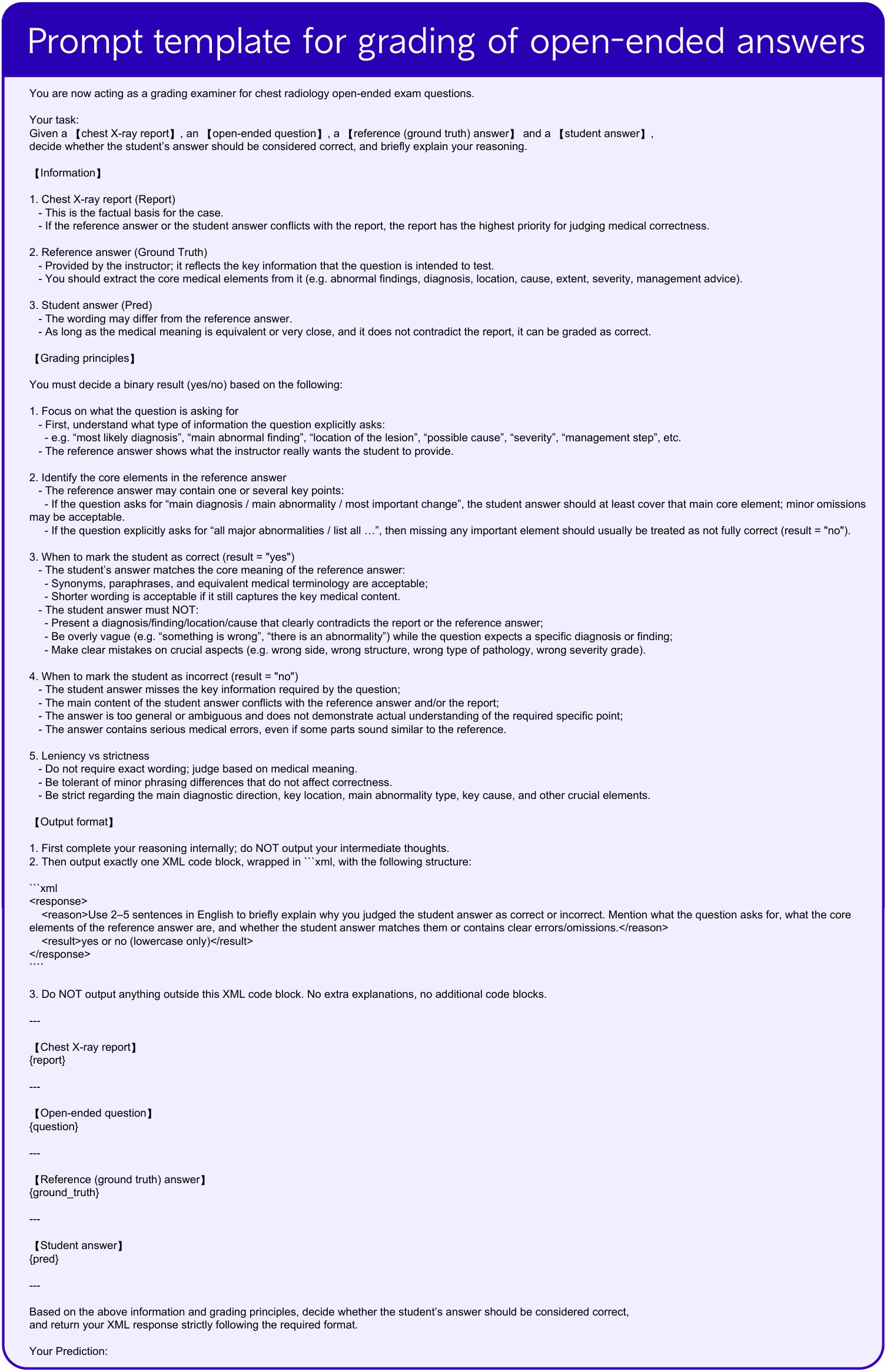}
{Prompt template used for LLM-as-a-Judge grading of open-ended answers. The judge determines whether the predicted answer is medically correct given the radiology report, the question, and the reference answer.}
[fig:judge_prompt]


\mideafullpagefigure[max width=0.96\paperwidth,max height=0.88\paperheight]{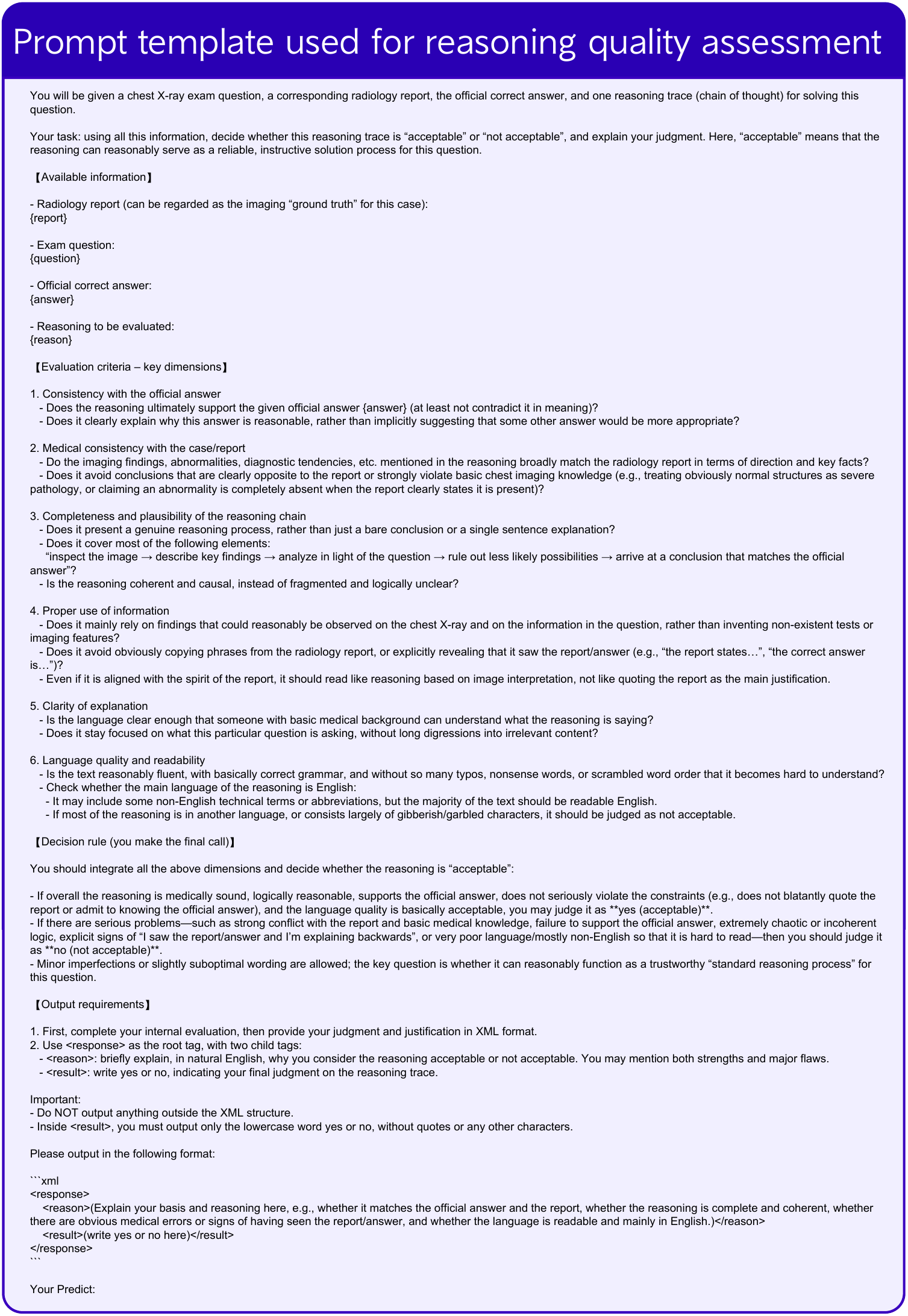}
{Prompt template used for reasoning quality assessment in auxiliary reward construction. The judge determines whether a reasoning trace can serve as an acceptable and reliable solution process for the given chest X-ray question.}
[fig:reward_judge_prompt]


\subsection{Reasoning Quality Assessment for Auxiliary Reward}
\label{app:judge_reasoning_quality}

In the comparative experiments, we additionally consider a \textbf{Judge Reward} baseline, in which an LLM evaluates whether a single reasoning trace is acceptable and converts this judgment into an auxiliary reward signal. Specifically, given a radiology report, an exam question, the official correct answer, and one reasoning trace, the judge determines whether the reasoning can serve as a medically sound, logically coherent, and pedagogically reliable solution process for the question. The corresponding prompt is shown in Figure~\ref{fig:reward_judge_prompt}.

The judge returns a binary XML decision with \texttt{<result>} equal to either \texttt{yes} or \texttt{no}. We map this decision to
\begin{equation}
s_i^{\mathrm{judge}}\in\{0,1\},
\end{equation}
where \texttt{yes} corresponds to $s_i^{\mathrm{judge}}=1$ and \texttt{no} corresponds to $s_i^{\mathrm{judge}}=0$. This binary signal is then used to construct the reward in the judge-reward baseline:
\begin{equation}
\tilde r_i^{\mathrm{rea}}
=
m_i^{\mathrm{fmt}}
\bigl(1+s_i^{\mathrm{judge}}+\mathbf{1}[\hat A_i=A_i^*]\bigr).
\end{equation}

The prompt asks the judge to evaluate the reasoning in terms of answer consistency, medical consistency with the report, coherence and plausibility of the reasoning process, proper use of available information, clarity of explanation, and overall readability. In essence, a reasoning trace is accepted only when it is broadly consistent with both the report and the official answer, follows a coherent and medically plausible reasoning process, avoids obvious factual conflicts or unsupported speculation, and remains readable.

As in training-time open-ended answer judging, malformed XML outputs or invalid \texttt{<result>} labels are handled by the same fallback rule described above. Each reasoning trace is judged once in this setting.

\section{Training Details}
\label{app:training_details}

\subsection{Training Setup}
\label{app:training_setup}

We optimize both the Reasoner and the Teacher with GRPO under the staged training scheme described in Section~\ref{sec:method_training_procedure}. Unless otherwise specified, all compared methods share the same GRPO training pipeline and hyperparameter configuration, and differ only in the reward/objective design. All experiments are conducted on two compute nodes, each equipped with 8 NVIDIA L40 GPUs (48GB): one node is dedicated to RL training, and the other serves as a unified reward-service node.

The training node is used for model optimization. The reward-service node hosts one Teacher instance for reference CoT generation and multiple \texttt{Qwen3-4B-Instruct-2507} judge instances for pairwise CoT comparison and open-ended answer judging. Reward requests are managed through a unified Ray-based service with batched scheduling and inference. During RL training, we use the GRPO objective together with KL regularization against the reference policy. The main reproduction-relevant hyperparameters are summarized in Table~\ref{tab:training_hyperparameters}.

For reference CoT generation, the frozen Teacher at each stage, denoted by $\tilde{\mathcal T}_i$, is served as a separate inference instance on the reward-service node. For each training sample, it generates $K=10$ reference CoTs using nucleus sampling with temperature $0.7$ and top-$p=0.95$. We set the maximum generation length to 4096 tokens. For image preprocessing, we use a pixel range of $[262144, 524288]$ to control the visual input resolution. These settings are shared across Teacher reference generation in all T2R stages.

\begin{table*}[h!]
\centering
\small
\caption{Key hyperparameters used in staged GRPO training. Unless otherwise specified, these settings are shared across the Teacher, the Reasoner, and all comparison methods; differences arise only from the reward/objective design.}
\label{tab:training_hyperparameters}
\begin{tabular}{lll}
\toprule
\textbf{Category} & \textbf{Setting} & \textbf{Value} \\
\midrule
\multirow{3}{*}{Algorithm}
& Objective & GRPO \\
& KL regularization & enabled \\
& KL coefficient & $1\times 10^{-2}$ \\
\midrule

\multirow{3}{*}{Rollout}
& Rollouts per prompt & 16 \\
& Temperature & 1.0 \\
& Top-$p$ & 1.0 \\
\midrule

\multirow{3}{*}{Data}
& Train rollout batch size & 128 \\
& Max prompt length & 5120 \\
& Max response length & 1024 \\
\midrule

\multirow{8}{*}{Optimization}
& Learning rate & $1\times 10^{-6}$ \\
& Weight decay & $1\times 10^{-2}$ \\
& Optimizer & AdamW (bf16) \\
& Max gradient norm & 1.0 \\
& Gradient checkpointing & enabled \\
& Vision tower freezing & disabled \\
& Padding-free training & enabled \\
& Dynamic batching & enabled \\
\midrule

Training
& Epochs per stage/split & 1 \\
\midrule

\multirow{5}{*}{System}
& RL training node & 1 node, 8 $\times$ NVIDIA L40 (48GB) \\
& Reward-service node & 1 node, 8 $\times$ NVIDIA L40 (48GB) \\
& \quad Teacher reference instance & 1 GPU instance \\
& \quad Judge instances & 7 GPU instances of \texttt{Qwen3-4B-Instruct-2507} \\
& Reward serving & unified Ray-based scheduling and request dispatch \\
\bottomrule
\end{tabular}
\end{table*}

\subsection{Reasoner Prompt}
\label{app:prompt_construction}

The Reasoner prompt is instantiated from three placeholders: \texttt{\{question\}}, \texttt{\{question\_type\}}, and \texttt{\{question\_type\_desc\}}. Here, \texttt{\{question\}} denotes the exam question, while \texttt{\{question\_type\}} and \texttt{\{question\_type\_desc\}} specify the question type and the corresponding answer-format instruction.

For open-ended questions, \texttt{\{question\_type\}} is set to \texttt{Open Ended Question}, and the answer-format instruction requires a short free-text answer. For single-choice questions, \texttt{\{question\_type\}} is set to \texttt{Single Choice Question}, and the answer-format instruction requires the model to output only the letter of the correct option, without brackets, quotation marks, or additional text.

\begin{figure*}[h!]
    \centering
    \includegraphics[width=\textwidth]{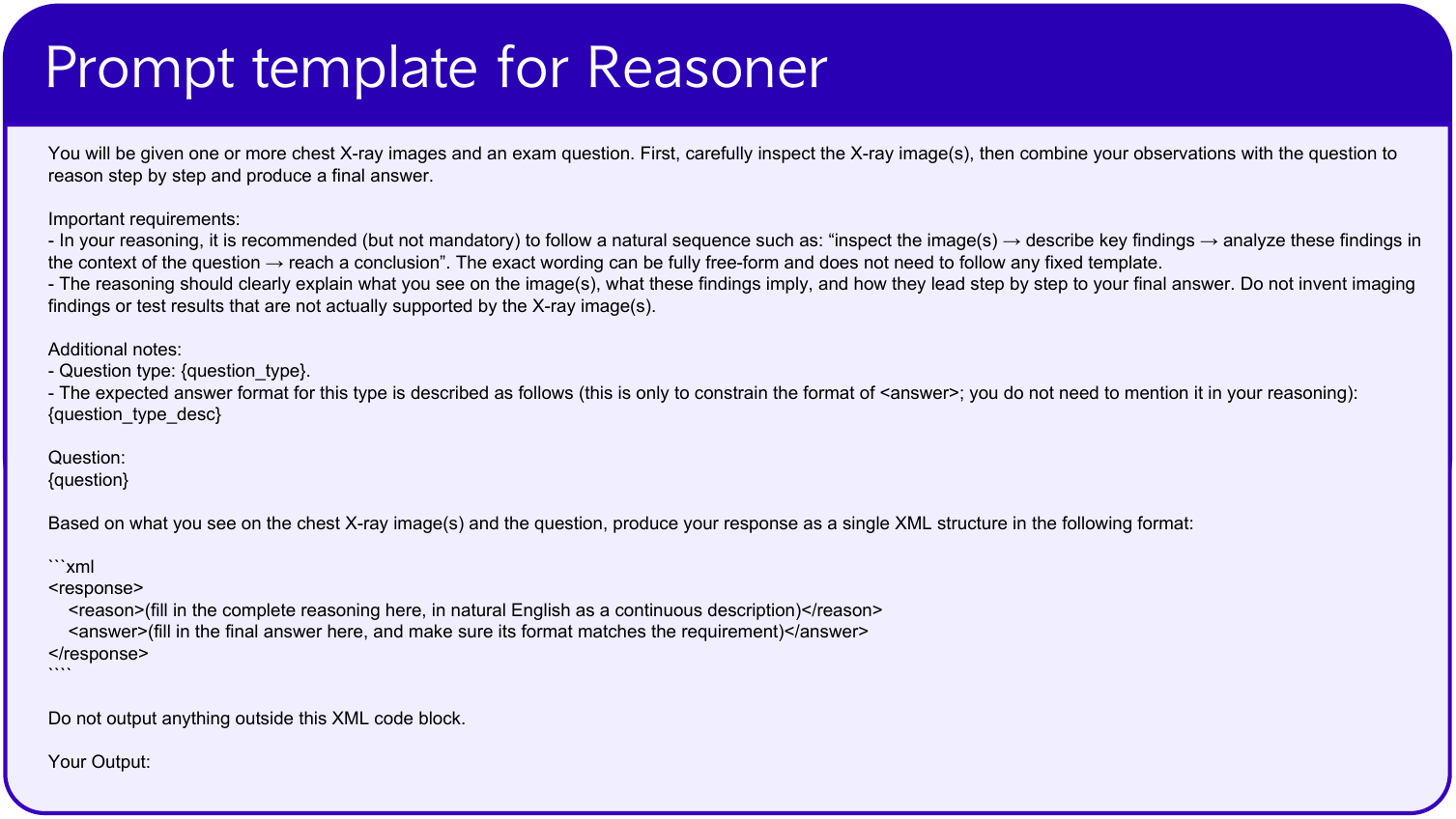}
    \caption{Prompt template used for Reasoner training. The prompt includes the exam question, the question type, and the corresponding answer-format instruction, and requires the model to return a single XML response containing both \texttt{<reason>} and \texttt{<answer>}.}
    \label{fig:reasoner_prompt}
\end{figure*}

These fields are inserted into a unified prompt template that requires the model to return a single XML response containing both a \texttt{<reason>} field and an \texttt{<answer>} field. The full prompt template is shown in Figure~\ref{fig:reasoner_prompt}.

\subsection{Teacher Prompt}
\label{app:teacher_prompt}

The Teacher prompt is instantiated from five placeholders: \texttt{\{question\}}, \texttt{\{question\_type\}}, \texttt{\{question\_type\_desc\}}, \texttt{\{answer\}}, and \texttt{\{report\}}. Here, \texttt{\{question\}} denotes the exam question, \texttt{\{question\_type\}} and \texttt{\{question\_type\_desc\}} specify the question type and the corresponding answer-format instruction, \texttt{\{answer\}} is the official correct answer, and \texttt{\{report\}} is the radiology report for the same case.

Unlike the Reasoner, the Teacher is prompted to reconstruct a medically sound reasoning process while using the radiology report and the official answer only as alignment signals. The generated reasoning must be written as if the model had access only to the chest X-ray image and the exam question. Accordingly, the Teacher is explicitly instructed not to mention the radiology report, not to quote the official answer, and not to present the reasoning as a reverse explanation of a known conclusion.

As in the Reasoner prompt, the question type and the corresponding answer-format instruction are included to align the generated reasoning with the downstream task setting. For open-ended questions, the prompt specifies a short free-text answer format; for single-choice questions, it specifies that the correct option should be represented by its letter only. These instructions are used only for task alignment and should not be explicitly referenced in the generated reasoning.

\begin{figure*}[h!]
    \centering
    \includegraphics[width=\textwidth]{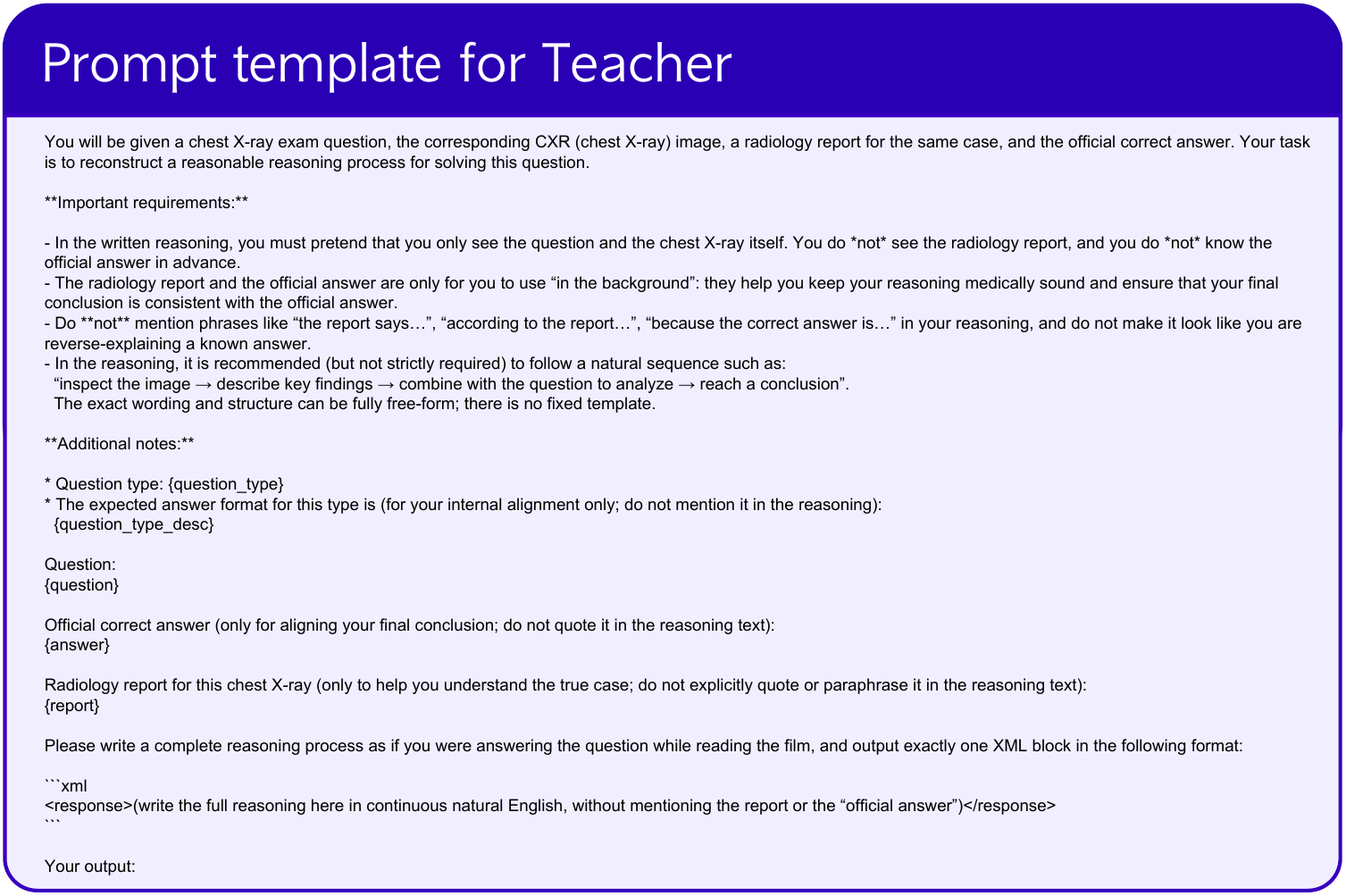}
    \caption{Prompt template used for Teacher training. The Teacher receives the exam question, question type, official correct answer, and radiology report, and is required to reconstruct a medically sound reasoning process while writing as if only the image and the question were observed. The output must be a single XML response containing a \texttt{<response>} field.}
    \label{fig:teacher_prompt}
\end{figure*}

The Teacher is required to return a single XML response containing one \texttt{<response>} field, which stores the full reasoning trace in continuous natural English. The full prompt template is shown in Figure~\ref{fig:teacher_prompt}.

\section{Detailed Derivations for the Reasoner Objective}
\label{app:reasoner_derivations}

\subsection{Non-Degenerate Case}
\label{app:nondegenerate_derivation}

In the non-degenerate case, the base reward already induces a non-trivial group-wise preference through
\begin{equation}
a_i^{\mathrm{raw}}
=
\tilde r_i^{\mathrm{rea}}
-\frac{1}{G}\sum_{j=1}^{G}\tilde r_j^{\mathrm{rea}},
\end{equation}
which defines the subsets
\begin{equation}
\mathcal P^+=\{i\in\mathcal I: a_i^{\mathrm{raw}}>0\},
\qquad
\mathcal P^-=\{i\in\mathcal I: a_i^{\mathrm{raw}}<0\}.
\end{equation}
In this case, the competition bonus is introduced only to refine the relative ordering within each subset; it should not alter the positive/negative partition induced by $a_i^{\mathrm{raw}}$. We therefore verify that the proposed scaling preserves the sign of the base-reward advantage.

For $i\in\mathcal P^+$ and $i\in\mathcal P^-$, we define the subset-wise standardized patterns
\begin{equation}
p_i^+
=
\frac{
s_i^{\mathrm{rea}}-\mu_+^{\mathrm{rea}}
}{
\sigma_+^{\mathrm{rea}}
},
\qquad
p_i^-
=
\frac{
s_i^{\mathrm{rea}}-\mu_-^{\mathrm{rea}}
}{
\sigma_-^{\mathrm{rea}}
}.
\end{equation}
We assume that the corresponding within-subset standard deviation is nonzero; when it vanishes, the competition bonus on that subset is set to zero.

The competition bonus is
\begin{equation}
\Delta_i^{\mathrm{rea}}
=
\begin{cases}
\eta_+^{\mathrm{rea}}\, p_i^+, & i\in\mathcal P^+,\\[2pt]
\eta_-^{\mathrm{rea}}\, p_i^-, & i\in\mathcal P^-,\\[2pt]
0, & i\notin \mathcal P^+\cup\mathcal P^-,
\end{cases}
\end{equation}
and the final advantage is
\begin{equation}
a_i^{\mathrm{final}}=a_i^{\mathrm{raw}}+\Delta_i^{\mathrm{rea}}.
\end{equation}

To preserve the sign induced by the base reward, the scaling coefficients are defined as
\begin{equation}
\eta_+^{\mathrm{rea}}
=
\alpha
\min_{i\in\mathcal P^+,\, p_i^+<0}
\frac{a_i^{\mathrm{raw}}}{-p_i^+},
\qquad
\eta_-^{\mathrm{rea}}
=
\alpha
\min_{i\in\mathcal P^-,\, p_i^->0}
\frac{-a_i^{\mathrm{raw}}}{p_i^-},
\end{equation}
where $\alpha\in(0,1)$. If the constraining set $\{i\in\mathcal P^+:p_i^+<0\}$ is empty, we set $\eta_+^{\mathrm{rea}}=0$; similarly, if $\{i\in\mathcal P^-:p_i^->0\}$ is empty, we set $\eta_-^{\mathrm{rea}}=0$.

We now verify that this construction preserves the sign of $a_i^{\mathrm{raw}}$.

\paragraph{Positive subset.}
Let $i\in\mathcal P^+$. Then $a_i^{\mathrm{raw}}>0$. If $p_i^+\ge 0$, we have
\begin{equation}
a_i^{\mathrm{final}}
=
a_i^{\mathrm{raw}}+\eta_+^{\mathrm{rea}}p_i^+
\ge
a_i^{\mathrm{raw}}
>
0,
\end{equation}
so the sign is preserved automatically.

It remains to consider the constraining case $p_i^+<0$. Since
\begin{equation}
\min_{j\in\mathcal P^+,\, p_j^+<0}
\frac{a_j^{\mathrm{raw}}}{-p_j^+}
\le
\frac{a_i^{\mathrm{raw}}}{-p_i^+},
\end{equation}
it follows that
\begin{equation}
\eta_+^{\mathrm{rea}}
=
\alpha
\min_{j\in\mathcal P^+,\, p_j^+<0}
\frac{a_j^{\mathrm{raw}}}{-p_j^+}
\le
\alpha\frac{a_i^{\mathrm{raw}}}{-p_i^+}.
\end{equation}
Because $p_i^+<0$, multiplying both sides by $p_i^+$ reverses the inequality:
\begin{equation}
\eta_+^{\mathrm{rea}}p_i^+
\ge
-\alpha a_i^{\mathrm{raw}}.
\end{equation}
Hence,
\begin{equation}
a_i^{\mathrm{final}}
=
a_i^{\mathrm{raw}}+\eta_+^{\mathrm{rea}}p_i^+
\ge
(1-\alpha)a_i^{\mathrm{raw}}
>
0,
\end{equation}
where the strict inequality follows from $\alpha\in(0,1)$ and $a_i^{\mathrm{raw}}>0$.

\paragraph{Negative subset.}
Let $i\in\mathcal P^-$. Then $a_i^{\mathrm{raw}}<0$. If $p_i^-\le 0$, we have
\begin{equation}
a_i^{\mathrm{final}}
=
a_i^{\mathrm{raw}}+\eta_-^{\mathrm{rea}}p_i^-
\le
a_i^{\mathrm{raw}}
<
0,
\end{equation}
so the sign is again preserved automatically.

Now consider the constraining case $p_i^->0$. Since
\begin{equation}
\min_{j\in\mathcal P^-,\, p_j^->0}
\frac{-a_j^{\mathrm{raw}}}{p_j^-}
\le
\frac{-a_i^{\mathrm{raw}}}{p_i^-},
\end{equation}
we obtain
\begin{equation}
\eta_-^{\mathrm{rea}}
=
\alpha
\min_{j\in\mathcal P^-,\, p_j^->0}
\frac{-a_j^{\mathrm{raw}}}{p_j^-}
\le
\alpha\frac{-a_i^{\mathrm{raw}}}{p_i^-}.
\end{equation}
Since $p_i^->0$, multiplying by $p_i^-$ preserves the inequality:
\begin{equation}
\eta_-^{\mathrm{rea}}p_i^-
\le
-\alpha a_i^{\mathrm{raw}}.
\end{equation}
Therefore,
\begin{equation}
a_i^{\mathrm{final}}
=
a_i^{\mathrm{raw}}+\eta_-^{\mathrm{rea}}p_i^-
\le
(1-\alpha)a_i^{\mathrm{raw}}
<
0,
\end{equation}
again because $\alpha\in(0,1)$ and $a_i^{\mathrm{raw}}<0$.

Combining the two subset analyses, we conclude that the competition bonus refines the within-subset ordering without changing the sign induced by the base reward. In particular, $a_i^{\mathrm{final}}>0$ for all $i\in\mathcal P^+$, $a_i^{\mathrm{final}}<0$ for all $i\in\mathcal P^-$, and $a_i^{\mathrm{final}}=0$ for all $i\notin\mathcal P^+\cup\mathcal P^-$. Thus, the proposed bonus is sign-preserving and cannot alter the positive/negative assignment induced by the base reward.

\subsection{Degenerate Case}
\label{app:degenerate_derivation}

In the degenerate case, all samples in the group receive the same base reward, so
\begin{equation}
a_i^{\mathrm{raw}}=0,\qquad \forall i\in\mathcal I.
\end{equation}
Hence, the base reward induces no non-trivial group-wise preference, and the final advantage must be reconstructed directly from the competition scores.

For a candidate threshold $\tau$, we define
\begin{equation}
\mathcal P^-(\tau)=\{i\in\mathcal I:\, s_i^{\mathrm{rea}}\le \tau\},
\qquad
\mathcal P^+(\tau)=\{i\in\mathcal I:\, s_i^{\mathrm{rea}}>\tau\}.
\end{equation}
We consider candidate thresholds drawn from the ordered values of $\{s_i^{\mathrm{rea}}\}_{i=1}^{G}$. Among all thresholds that induce non-empty subsets on both sides, we choose the largest feasible $\tau$ satisfying
\begin{equation}
\sum_{i\in\mathcal P^-(\tau)} s_i^{\mathrm{rea}}
<
\gamma
\sum_{i\in\mathcal P^+(\tau)} s_i^{\mathrm{rea}},
\end{equation}
where $\gamma>0$ is a hyperparameter. If no feasible threshold exists, we set
\begin{equation}
a_i^{\mathrm{final}}=0,\qquad \forall i\in\mathcal I.
\end{equation}
We therefore consider only the feasible case, which yields induced subsets $\mathcal P^+$ and $\mathcal P^-$.

We first globally center the competition scores:
\begin{equation}
p_i=s_i^{\mathrm{rea}}-\mu^{\mathrm{rea}},
\qquad
\mu^{\mathrm{rea}}=\frac{1}{G}\sum_{j=1}^{G}s_j^{\mathrm{rea}}.
\end{equation}
By construction,
\begin{equation}
\sum_{i=1}^{G} p_i = 0.
\end{equation}

To obtain a reconstructed signal that is both zero-mean and sign-consistent with the induced partition, we shift only the positive subset:
\begin{equation}
\hat p_i=
\begin{cases}
p_i+\lambda, & i\in\mathcal P^+,\\[2pt]
p_i, & i\in\mathcal P^-.
\end{cases}
\end{equation}
Let
\begin{equation}
m=|\mathcal P^+|,
\qquad
n=|\mathcal P^-|,
\qquad
m+n=G.
\end{equation}
The group mean of $\{\hat p_i\}_{i=1}^G$ is then
\begin{equation}
\frac{1}{G}\sum_{i=1}^{G}\hat p_i
=
\frac{1}{G}\left(\sum_{i=1}^{G}p_i + m\lambda\right)
=
\frac{m\lambda}{G}.
\end{equation}

The final advantage is defined as
\begin{equation}
a_i^{\mathrm{final}}
=
\hat p_i-\frac{1}{G}\sum_{j=1}^{G}\hat p_j.
\end{equation}
To ensure consistency with the induced partition, we require
\begin{equation}
a_i^{\mathrm{final}}\ge 0,\qquad \forall i\in\mathcal P^+,
\end{equation}
and
\begin{equation}
a_i^{\mathrm{final}}\le 0,\qquad \forall i\in\mathcal P^-.
\end{equation}

\paragraph{Positive subset constraint.}
For $i\in\mathcal P^+$,
\begin{equation}
a_i^{\mathrm{final}}
=
\hat p_i-\frac{1}{G}\sum_{j=1}^{G}\hat p_j
=
p_i+\lambda-\frac{m\lambda}{G}
=
p_i+\frac{n}{G}\lambda.
\end{equation}
Hence, the non-negativity constraint reduces to
\begin{equation}
p_i+\frac{n}{G}\lambda\ge 0
\quad\Longleftrightarrow\quad
\lambda \ge -\frac{G}{n}p_i.
\end{equation}
To satisfy this for all $i\in\mathcal P^+$, it is sufficient to require
\begin{equation}
\lambda \ge -\frac{G}{n}\min_{i\in\mathcal P^+}p_i.
\end{equation}

\paragraph{Negative subset constraint.}
For $i\in\mathcal P^-$,
\begin{equation}
a_i^{\mathrm{final}}
=
\hat p_i-\frac{1}{G}\sum_{j=1}^{G}\hat p_j
=
p_i-\frac{m\lambda}{G}.
\end{equation}
Hence, the non-positivity constraint reduces to
\begin{equation}
p_i-\frac{m\lambda}{G}\le 0
\quad\Longleftrightarrow\quad
\lambda \ge \frac{G}{m}p_i.
\end{equation}
To satisfy this for all $i\in\mathcal P^-$, it is sufficient to require
\begin{equation}
\lambda \ge \frac{G}{m}\max_{i\in\mathcal P^-}p_i.
\end{equation}

Combining the two subset constraints, the minimum feasible shift is
\begin{equation}
\lambda_{\min}
=
\max\!\left(
-\frac{G}{|\mathcal P^-|}\min_{i\in\mathcal P^+} p_i,\;
\frac{G}{|\mathcal P^+|}\max_{i\in\mathcal P^-} p_i
\right),
\end{equation}
and we set
\begin{equation}
\lambda=\max(\lambda_{\min},0).
\end{equation}
With this choice of $\lambda$, the reconstructed final advantage is zero-mean at the group level and sign-consistent with the partition induced by the competition scores. It therefore provides a valid replacement training signal when the original group-level advantage collapses to zero.
\section{Qualitative Case Studies}
\label{app:qualitative_cases}

We provide qualitative examples in Figure~\ref{fig:cot_case1}--\ref{fig:cot_case3} to compare the reasoning traces generated by \textbf{RLVR Only}, \textbf{Judge Reward}, \textbf{Frozen Teacher}, and \textbf{T2R-R3}. Across these examples, different supervision signals are associated with visibly different CoT structures. \textbf{RLVR Only} tends to produce answer-oriented rationales with limited intermediate analysis, which is consistent with its lack of explicit CoT-level supervision. \textbf{Judge Reward} improves local fluency and coherence, but the resulting reasoning often remains relatively brief and stops after identifying a few relevant findings. \textbf{Frozen Teacher} further encourages more organized rationales, including explicit summaries of key findings. In contrast, \textbf{T2R-R3} more consistently follows a complete reasoning pattern: it first analyzes the chest X-ray findings, then summarizes the relevant evidence, evaluates the candidate options when applicable, and finally derives the answer from the accumulated observations.

This qualitative pattern is consistent with the training signal provided by T2R. Directly judging a single CoT with an LLM can encourage the model to satisfy the predefined acceptability criteria in the judge prompt (Figure~\ref{fig:reward_judge_prompt}), but such supervision may become less informative as the model adapts to these criteria. By contrast, T2R relies on pairwise comparison against reference CoTs produced by a Teacher that is itself improved through self-competition. As the reference CoTs become stronger, the Reasoner continues to receive relative preference signals that favor more grounded, structured, and option-aware reasoning. These examples therefore suggest that comparison-based supervision can improve not only final answer prediction, but also the organization and reliability of the generated CoT.

\section{Limitations and Future Work}
\label{app:limitations_future_work}

T2R inherits several limitations from its reliance on LLM-as-a-Judge supervision. First, the quality of the comparison signal depends on the judge model having sufficient instruction-following ability and reliable reasoning capability. If the judge produces biased or inaccurate comparisons, the Reasoner may be optimized toward suboptimal preferences. Second, T2R introduces additional computational cost. To compute competition scores, the Teacher first generates $K$ reference CoTs for each Reasoner output, and an LLM judge then performs pairwise comparisons between the generated CoTs. This requires substantially more inference than standard RLVR or direct reward-based training. In our implementation, we mitigate this overhead by deploying a separate reward-service node and using Ray-based request dispatch, but improving the efficiency of comparison-based training remains an important direction.

Another limitation is that our current comparison prompt aggregates multiple criteria into a single overall judgment of which CoT is better. While this design is simple and effective, real-world medical reasoning often involves multiple distinct requirements, such as evidence grounding, answer support, logical consistency, restraint, and task-specific clinical preferences. A promising future direction is therefore to decompose the unified comparison into multiple finer-grained comparison dimensions, each associated with a specific rule or requirement and potentially assigned a different weight. This could make the supervision signal more explicit, controllable, and better aligned with practical clinical needs.

Finally, this work applies T2R only to CoT optimization for CXR VQA. However, the underlying idea is more general: comparison-based supervision can in principle be applied to other model outputs, such as solution strategies, planning traces, code, or other structured responses. Extending T2R beyond medical VQA and designing task-specific comparison schemes for broader forms of model output are promising directions for future work.
\section{Broader Impacts}
\label{app:broader_impacts}

This work studies reinforcement-learning-based reasoning optimization for chest X-ray visual question answering. Its potential positive impact is to improve the reliability and interpretability of medical multimodal models by encouraging more evidence-grounded and coherent reasoning traces. Such techniques may be useful for developing decision-support systems, medical education tools, or research assistants that provide more transparent intermediate reasoning.
At the same time, this work also has potential risks. Models trained with T2R may still produce incorrect answers or medically implausible rationales, and LLM-as-a-Judge supervision may introduce biases or failure modes from the judge model. In clinical settings, incorrect or overconfident reasoning could negatively affect decision-making if used without expert oversight. Therefore, the models and methods studied in this paper should not be used as standalone clinical decision systems. Any practical deployment would require careful validation on diverse populations, monitoring for bias and failure cases, privacy-preserving data handling, and use under the supervision of qualified medical professionals.

\mideafullpagefigure[max width=0.96\paperwidth,max height=0.88\paperheight]{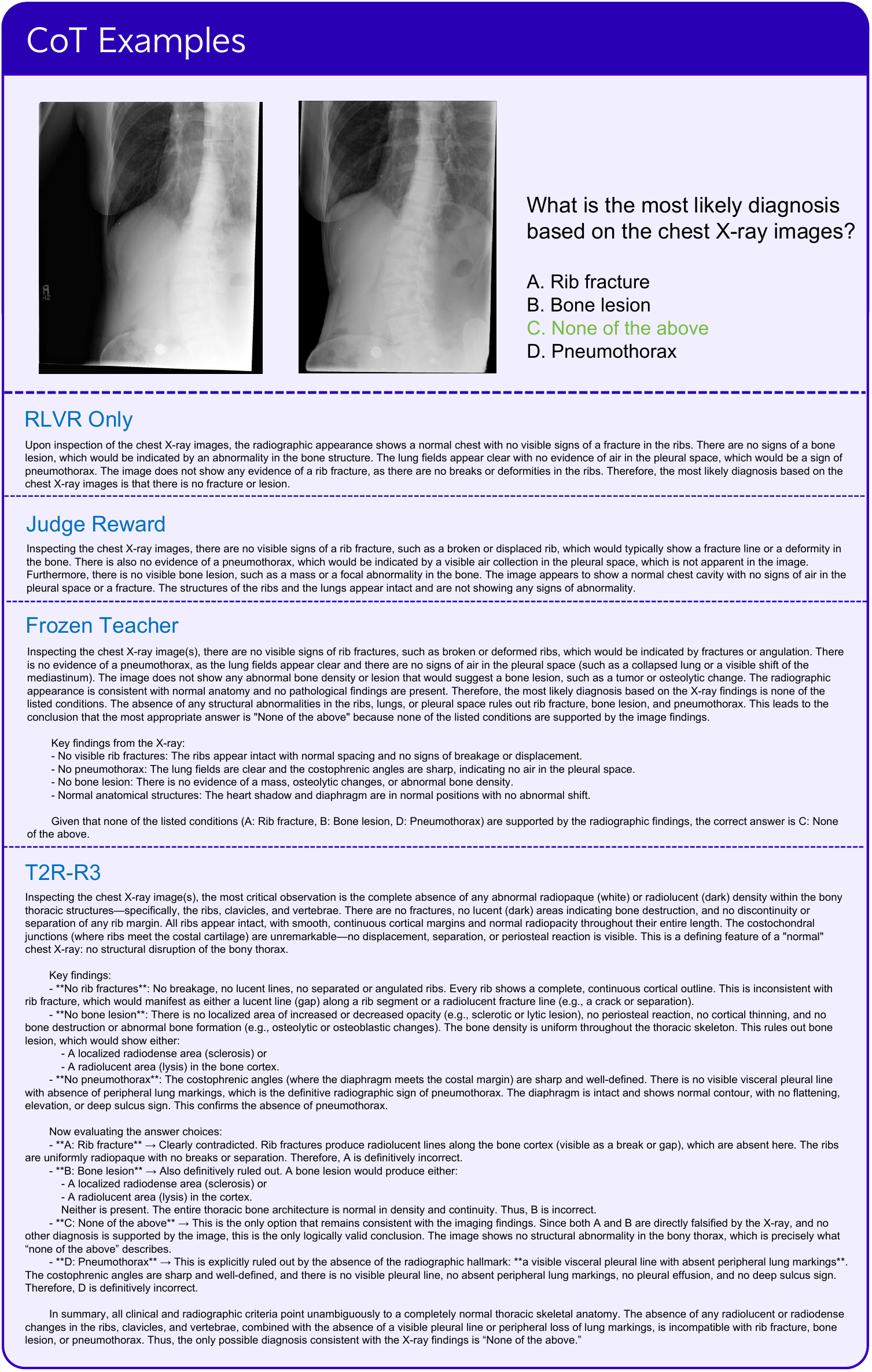}
{Qualitative comparison of CoTs generated by different models.}
[fig:cot_case1]


\mideafullpagefigure[max width=0.96\paperwidth,max height=0.88\paperheight]{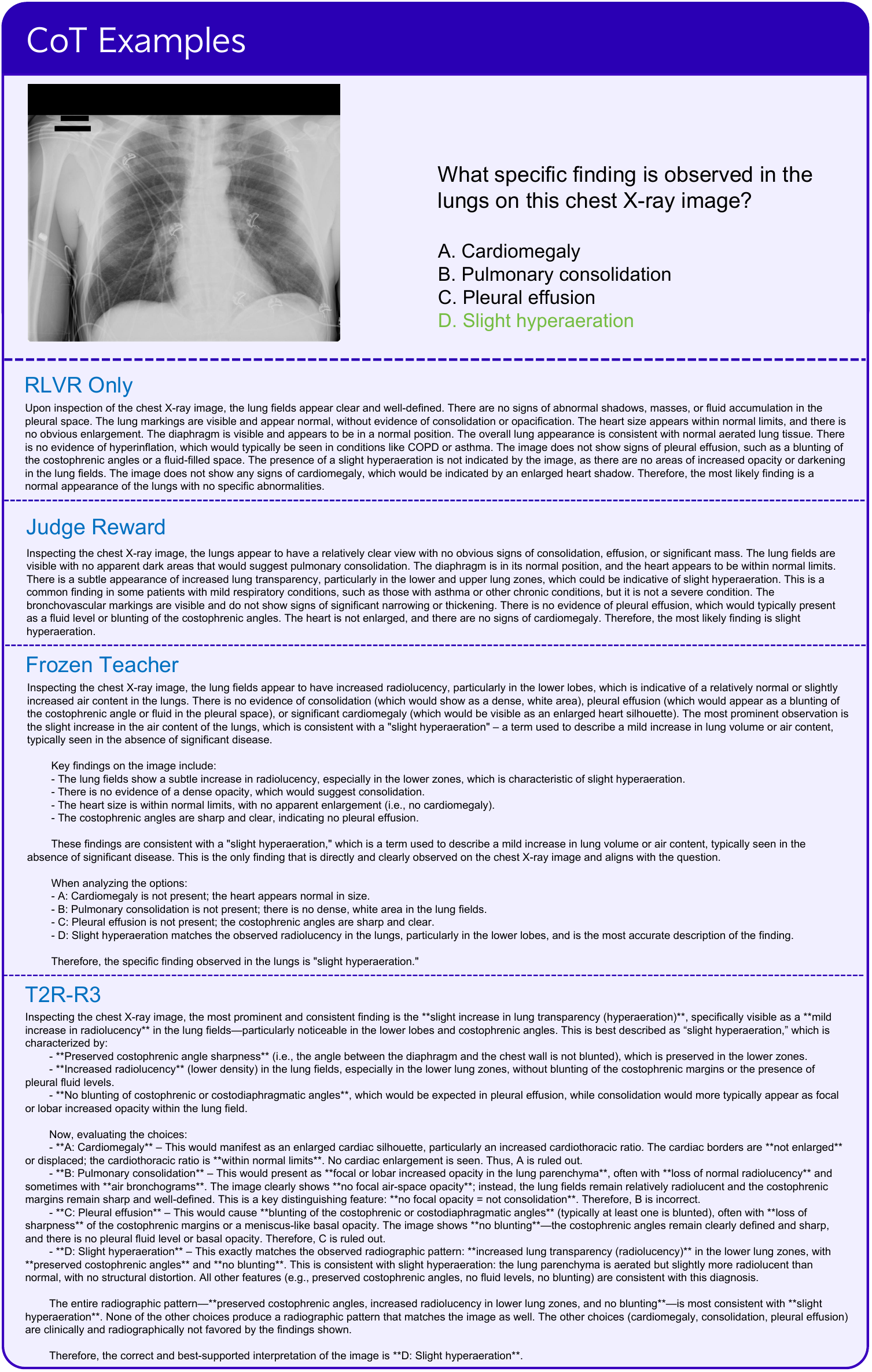}
{Qualitative comparison of CoTs generated by different models.}
[fig:cot_case2]


\mideafullpagefigure[max width=0.96\paperwidth,max height=0.88\paperheight]{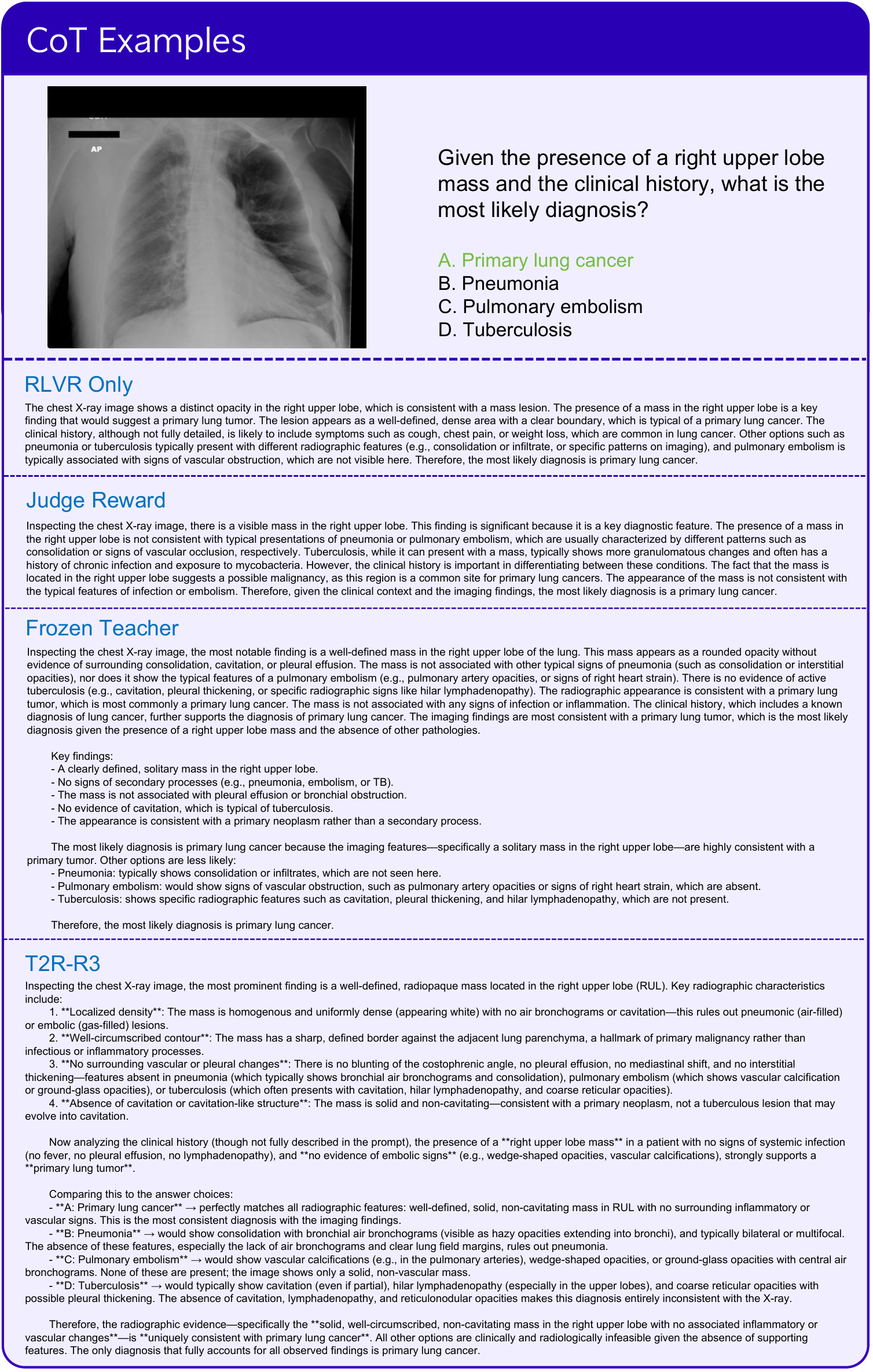}
{Qualitative comparison of CoTs generated by different models.}
[fig:cot_case3]



%

\end{document}